\documentclass[preprint,12pt]{elsarticle}

 \setcitestyle{authoryear,open={(},close={)}} 
\usepackage{xcolor}
\usepackage{amssymb}
\usepackage{graphicx}
\usepackage{amsmath}
\usepackage{blindtext}
\usepackage{comment}
\usepackage{ulem}
\usepackage{amsfonts}
\usepackage{mathtools}
\usepackage{booktabs}
\usepackage{multirow}
\usepackage{subcaption}
\usepackage[]{lineno}
\usepackage{hyperref}

\journal{}

\begin{document}

\begin{frontmatter}



\title{Fire-Image-DenseNet (FIDN) for predicting wildfire burnt area using remote sensing data}


\author[inst1,inst8]{Bo Pang}
\author[inst3]{Sibo Cheng\corref{cor1}}
\author[inst4]{Yuhan Huang}
\author[inst4]{Yufang Jin}
\author[inst2,inst6]{Yike Guo}
\author[inst5]{I. Colin Prentice}
\author[inst7]{Sandy P. Harrison}
\author[inst1,inst2]{Rossella Arcucci}

\affiliation[inst1]{
    organization={Department of Earth Science and Engineering, Imperial College London},
    city={London},
    country={UK}
}

\affiliation[inst2]{
    organization={Data Science Institute, Department of Computing, Imperial College London},
    city={London},
    country={UK}
}
\affiliation[inst3]{
    organization={CEREA, École des Ponts and EDF R\&D, Institut Polytechnique de Paris},
    city={Île-de-France},
    country={France}
}

\affiliation[inst4]{
    organization={Department of Land, Air and Water Resources, University of California, Davis},
    city={California},
    country={USA}
}

\affiliation[inst5]{
    organization={Georgina Mace Centre for the Living Planet, Department of Life Sciences, Imperial College London},
    city={London},
    country={UK}
}

\affiliation[inst7]{
    organization={Geography \& Environmental Science, University of Reading},
    city={Reading},
    country={UK}
}

\affiliation[inst6]{
    organization={ Department of Computer Science and Engineering, Hong Kong university of science and technology},
    city={HongKong},
    country={China}
}

\affiliation[inst8]{
    organization={Computer Network Information Center, Chinese Academy of Sciences },
    city={Beijing},
    country={China}
}

\cortext[cor1]{Corresponding: sibo.cheng@enpc.fr}


\begin{abstract}

Predicting the extent of massive wildfires once ignited is essential to reduce the subsequent socioeconomic losses and environmental damage, but challenging because of the complexity of fire behaviour. Existing physics-based models are limited in predicting large or long-duration wildfire events. Here, we develop a deep-learning-based predictive model, Fire-Image-DenseNet (FIDN), that uses spatial features derived from both near real-time and reanalysis data on the environmental and meteorological drivers of wildfire. We trained and tested this model using more than 300 individual wildfires that occurred between 2012 and 2019 in the western US. In contrast to existing models, the performance of FIDN does not degrade with fire size or duration. Furthermore, it predicts final burnt area accurately even in very heterogeneous landscapes in terms of fuel density and flammability. The FIDN model showed higher accuracy, with a mean squared error (MSE) about 82\% and 67\% lower than those of the predictive models based on cellular automata (CA) and the minimum travel time (MTT) approaches, respectively. Its structural similarity index measure (SSIM) averages 97\%, outperforming the CA and FlamMap MTT models by 6\% and 2\%, respectively. Additionally, FIDN is approximately three orders of magnitude faster than both CA and MTT models. The enhanced computational efficiency and accuracy advancements offer vital insights for strategic planning and resource allocation for firefighting operations.
\end{abstract}

\begin{graphicalabstract}
\begin{figure*}[!htb]
    \centering
    \makebox[\textwidth][c]{
    \includegraphics[width=1.2\textwidth]{GraphicalAbstract2.pdf}
    }
\end{figure*}
\end{graphicalabstract}


\begin{highlights}
\item We propose a deep learning model framework to predict the final burnt area of wildfires.

\item This model integrates remote sensng observations and meteorological inputs.

\item The model is trained and tested using recent wildfire data in the western US.

\item This model outperforms the state-of-the-art methods in both accuracy and efficiency.

\end{highlights}

\begin{keyword}
Deep learning \sep Wildfire prediction \sep Densenet \sep FlamMap \sep Cellular automata

\end{keyword}

\end{frontmatter}


\section{Introduction}
The frequency and intensity of large wildfires have increased in many parts of the world in recent years~\citep{dutta2016big, iglesias2022us, san2022advance}. Large wildfires have a significant impact on ecological resources~\citep{keeley2019fire, halofsky2020changing}, local and regional climate~\citep{baro2017biomass, stocker2021observing}, social infrastructure~\citep{thomas2017costs, fraser2022wildfire, varga2022wildfires} and human life and well-being~\citep{johnston2012estimated,bowman2017human,yu2020bushfires, chen2021mortality}. Significant resources are spent on firefighting, preventing and managing wildfires~\citep{wang2021economic, simon2022costs}. Predicting the spread and potential final extent of a given wildfire timely is important for disaster response and management~\citep{fairbrother2005predicting,taylor2013wildfire}, potentially including decisions about the allocation of firefighting resources and community evacuations.

Several types of models have been developed to simulate fire spread~\citep{sullivan2009wildland}, including empirical models~\citep{plourde1997new, guariso2002simulation} and physics-based models~\citep{alexandridis2008cellular, anderson1982modelling,burgan1984behave,mcarthur1967fire}. Empirical models are based on statistical relationship between environmental factors and fire behaviour~\citep{sullivan2007reviewb}. Physics-based models rely on physical principles, such as Rate of spread (ROS) modelling~\citep{johnson2001forest} or Huygens wavelet principle~\citep{anderson1982modelling}.
Among them, the Cellular Automata(CA) approach proposed by~\citep{alexandridis2008cellular} uses regular square meshes to simulate fire propagation along the grid to the neighbour cells. Each cell is categorised as non-combustible, combustible, burning, or burnt. At each time step, the transition from combustible to burning follows a probability distribution, which depends on plant densities, plant species, wind speed, and topographical slope. The CA model has been widely adopted to simulate individual wildfires in Greece~\citep{alexandridis2008cellular}, Portugal~\citep{freire2019using}, Italy~\citep{trucchia2020propagator} and US~\citep{zheng2017forest}.
Another state-of-art approach has been incorporated in the FlamMap software~\citep{finney2006overview}, developed by the US Forest Service.  It formulates the fire growth using the Minimum Travel Time (MTT) algorithm ~\citep{finney2002fire}, which calculates the minimum time for a fire to propagate between nodes in a two-dimensional network.

However, both CA and MTT approaches require the simulation of a large number of high-dimensional environmental and climatic variables, and thus could be considerably time-consuming~\citep{papadopoulos2011comparative}~\citep{jain2020review}. While these models are adept at making short-term predictions of fire spread, their accuracy degrades with time~\citep{hoffman2016evaluating} because they assume that fire spread takes place under constant meteorological conditions and because of the difficulty in incorporating phenomena such as fire-generated weather~\citep{fromm2022understanding}, transitions from surface to crown fires~\citep{weise2018surface}, and the role of spotting in generating new fire fronts~\citep{martin2016spotting} - all of which become more important as wildfires burn over a longer time. Moreover, many parameters in currently available models need to be adjusted for local conditions and thus for individual fires (e.g.~\citep{alessandri2021parameter,cheng2022data} ), a process both data demanding and time-consuming.

Machine learning (ML) approaches have been used to overcome some of these limitations. For example,~\citep{denham2018using} implemented the CA model on the Graphics Processing Unit (GPU) and used the Genetic Algorithm (GA) search strategy to adjust input parameter values to improve speed and accuracy. The work of~\citep{zheng2017forest} used the Extreme Learning Machine (ELM) to replace the diffusion strategy of the CA model. While these studies are limited to optimising the physical model,~\citep{cheng2022data,cheng2022parameter} introduced a data-driven methodology that relies on a combination of convolutional autoencoder and Long-Short-Term Memory (LSTM) techniques. This approach aimed to approximate the output of the CA model while achieving a remarkable 1000-fold increase in speed. The same surrogate model has also been employed in developing a generative AI to further decrease the offline computational cost~\citep{cheng2023generative}. However, this surrogate model needs to be retrained for different ecoregions. The work of ~\citep{shadrin2024wildfire} presents a neural network algorithm based on the MA-Net architecture, designed to predict the spread of a wildfire over the next five days, including the speed and direction of the fire. However, the algorithm cannot directly predict the final burned area of the fire.
Therefore, there is still a need for a generic, fast and accurate fire prediction model capable of assimilating spatial information from remote-sensing data to predict fire behaviour throughout its duration.

To achieve this, an important task involves extracting features from spatial remote sensing data. Convolutional Neural Network (CNN)-based approaches are the dominant learning methods for image processing~\citep{bouvrie2006notes}. Through the utilization of convolutional operations, CNNs mimic the functioning of the human visual system, resulting in a substantial reduction in the number of parameters required for training~\citep{bouvrie2006notes}. This attribute made CNNs particularly adept at capturing localized patterns. The Densely Connected Convolutional Network (Densenet)~\citep{huang2017densely}, which connects the input of the previous layer with the output feature map of the current layer directly, required even fewer parameters and training data compared to traditional convolutional networks~\citep{wang2020densenet,sellami2019robust,fujino2019evolutionary,zhang2018sparse}. These advantages suggested that DenseNet could be effectively used for feature extraction in the context of fire prediction modelling, particularly when the amount of training data is limited.

Transformer-based models, another type of deep learning model,  are highly advanced and effective in many applications. However, they typically require large volumes of data for training to achieve optimal performance. In our study, the availability of real-world wildfire data was limited, which poses a challenge for training such data-intensive models. This limitation influenced our decision to focus on CNN-based models, which is better suited for the data constraints we encountered.

In this paper, we propose a novel deep learning scheme, named Fire-Image-DenseNet (FIDN), to predict the final burnt area using initial fire spread data alongside vegetation and meteorological variables as inputs. These inputs include land cover type, real-time and reanalysis data on biomass,  tree and grass density, water bodies, wind speed and direction, precipitation, in addition to topography. Instead of predicting daily fire progression. our goal is to directly predict the final burnt area once a fire is ignited, which can facilitate advanced fire fighting resource allocation and provide guidance for overall fire fighting strategy.  The proposed model is tested using the data from recent massive wildfires in the western US, including California. Unlike existing autoregressive predictive models such as those with recurrent neural networks \cite{cheng2022data,li2021prediction,cheng2023generative}, our Fire-Image-DenseNet (FIDN) maintains higher accuracy regardless of fire size or duration, and it is adaptable to heterogeneous landscapes with varying fuel densities and flammability. Furthermore, unlike most of empirical models, FIDN does not require separate adjustments of parameters for different fires. Compared to the state-of-the-art CA and MTT approaches, FIDN yields significant advantages by
\begin{itemize}
    \item reducing the average computation time by 99.92\%;
    \item improving the structural similarity (SSIM) by 1.8\%;
    \item improving the peak signal-to-noise ratio (PSNR) by 6\%;
    \item reducing the mean square error (MSE) by 67.7\%;
 \end{itemize}

The rest of the paper is organized as follows. Section~\ref{section:3} introduces the model structure and the training method of FIDN. In Section~\ref{section:2}, we describe the structure, sources, and processing methods of the dataset used. Numerical results are presented and discussed in Section~\ref{section:4}. Concluding remarks are presented in Section~\ref{section:5}.

\section{Methodology}
\label{section:3}
In this section, we present the structure, training methods and evaluation metrics of the proposed FIDN model.

\subsection{Overall Research Framework}
In this study, we propose the Fire-Image-DenseNet (FIDN) model for accurate prediction of wildfire burnt areas utilizing advanced deep learning techniques. Our approach is driven by the critical need for effective and timely wildfire forecasting of where fire would stop in the absence of human intervention, which can significantly enhance preventive measures and resource allocation during wildfire incidents. The FIDN model combines the strengths of DenseNet architecture for feature extraction with a custom forecasting network designed to produce high-resolution predictions of final burnt areas. Through this methodology, we aim to leverage satellite observationsof the first three days of fire spread and relevant environmental data to improve the accuracy of predicting final spatially explicit areas burnt via machine learning. It addresses the challenges posed by the complexities of various input data dimensions, and ultimately contribute to better wildfire management practices.

In the FIDN model, the architecture is structured as an encoder-decoder framework, drawing inspiration from AutoEncoder designs~\citep{hinton2006reducing}. The encoder is responsible for extracting rich feature representations from the input wildfire images at the initial stage and environmental data, effectively compressing the information required for accurate prediction. This compression plays a crucial role in highlighting the most relevant features while minimizing noise and irrelevant details, a concept central to AutoEncoder methodologies. Following the encoding process, the decoder reconstructs the high-resolution prediction of the burnt areas, enabling precise spatial representation. By applying the encoder-decoder strategy, the FIDN model not only leverages the powerful feature extraction capabilities of Convolutional Neural Networks (CNNs) but also enhances the model's ability to learn complex mappings from input data to improve predictions.

\subsection{Densely Connected Convolutional Networks}

The CNNs is an important method in the field of image feature extraction that has been evolving for over 20 years~\citep{lecun1989backpropagation}. Compared with fully connected neural networks, it uses convolutional operations that are more appropriate for processing two-dimensional image information and significantly reduce the number of parameters~\citep{bouvrie2006notes}. During the long development history of CNNs, many classical models have been proposed. From the initial 5-layer LeNet5~\citep{lecun1998gradient} to the Residual Network (ResNet)~\citep{he2016deep} with over 100 layers, CNNs have been enhanced gradually to extract image features. In recent years, two main directions have been proposed to improve the effectiveness of CNN, either by increasing the depth of the network such as ResNet~\citep{he2016deep} or by extending the width of the network such as GoogleNet's Inception~\citep{szegedy2016rethinking}.

As the depth of the network increases, the problem of vanishing gradients~\citep{hanin2018neural, huang2017densely} has been noticed, leading to a degradation of network performance~\citep{hochreiter1998vanishing}. ResNet attempts to address performance degradation through residual learning. Residual learning is making shortcut connections between layers, which allows the stacked layers to learn directly from the input layers~\citep{he2016deep}. GoogleNet's Inception, in contrast, attempts to add multiple filters simultaneously, superimposing their outputs and allowing the network to choose the combination of parameters and filters to be learned.~\citep{szegedy2016rethinking}. In summary, the essential idea for solving the gradient disappearance problem is to create paths between the early and later layers.

The work of~\citep{huang2017densely} was inspired by this idea to propose a new structurally uncomplicated but effective convolutional neural network (CNN) -- Densely Connected Convolutional Networks (DenseNet). In contrast to ResNet which sums the output of the previous layer and its linear transform passes to the next layer, the most significant improvement in~\citep{huang2017densely} was the \textit{Dense connection} which allows the input of each layer to be derived from the output of all previous layers. This connection in DenseNet solves the problem of vanishing gradients by enabling the gradients to propagate more efficiently through the network. As a consequence, the Densenet structure requires fewer parameters and fewer training epochs~\citep{huang2017densely}.

\begin{figure}[!htb]
    \begin{center}
        \includegraphics[width=\textwidth]{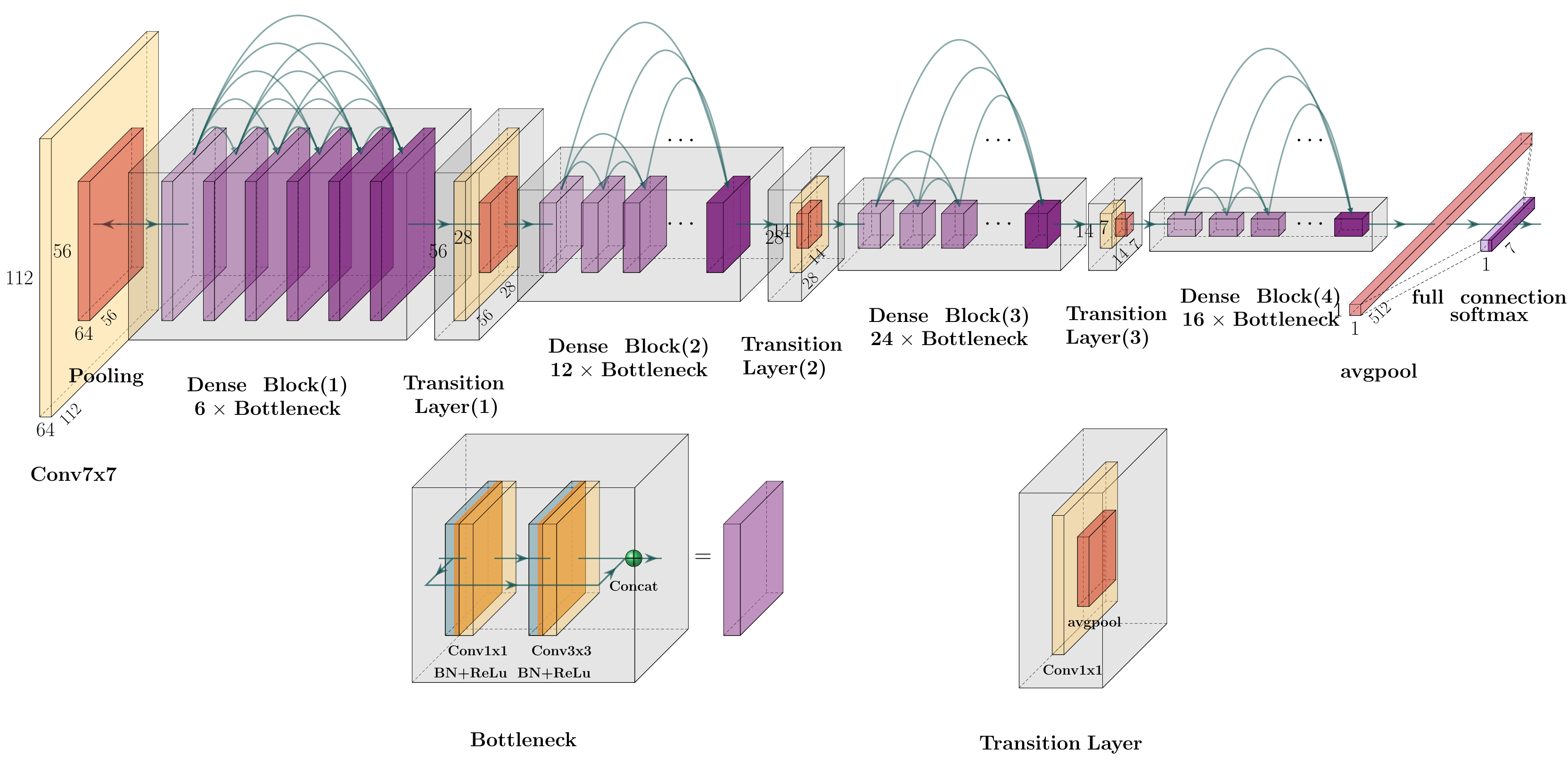}
    \end{center}
    \caption{\label{fig:densenet_structure}  DenseNet-121 architectures for ImageNet}
\end{figure}

More specifically, the DenseNet structure consists of two components:

\textbf{Dense Block}: For simplicity, the combination of a Batch Normalization(BN) layer, a ReLU Activation layer and a convolution layer is referred as \textit{Conv Block}; while the combination of a $1\times1$ Conv Block and a $3\times3$ Conv Block is called a \textit{Bottleneck layer}~\citep{huang2017densely}. Each \textit{Dense Block} is constructed by several Bottleneck layers, as shown in Figure~\ref{fig:densenet_structure}.

\textbf{Transition Layer}: It includes a 1$\times$1 Conv Block and an average pool layer which are used for dimensionality reduction~\citep{huang2017densely}.

Overall, DenseNet is composed of a convolutional layer, a pooling layer, four Dense Blocks interspersed with three Transition Layers and a fully connected layer for image classification. The complete structure diagram is shown in Figure~\ref{fig:densenet_structure}.

\subsection{Fire-Image-DenseNet(FIDN)}

\begin{figure}[!htb]
    \begin{center}
        \includegraphics[width=1\textwidth]{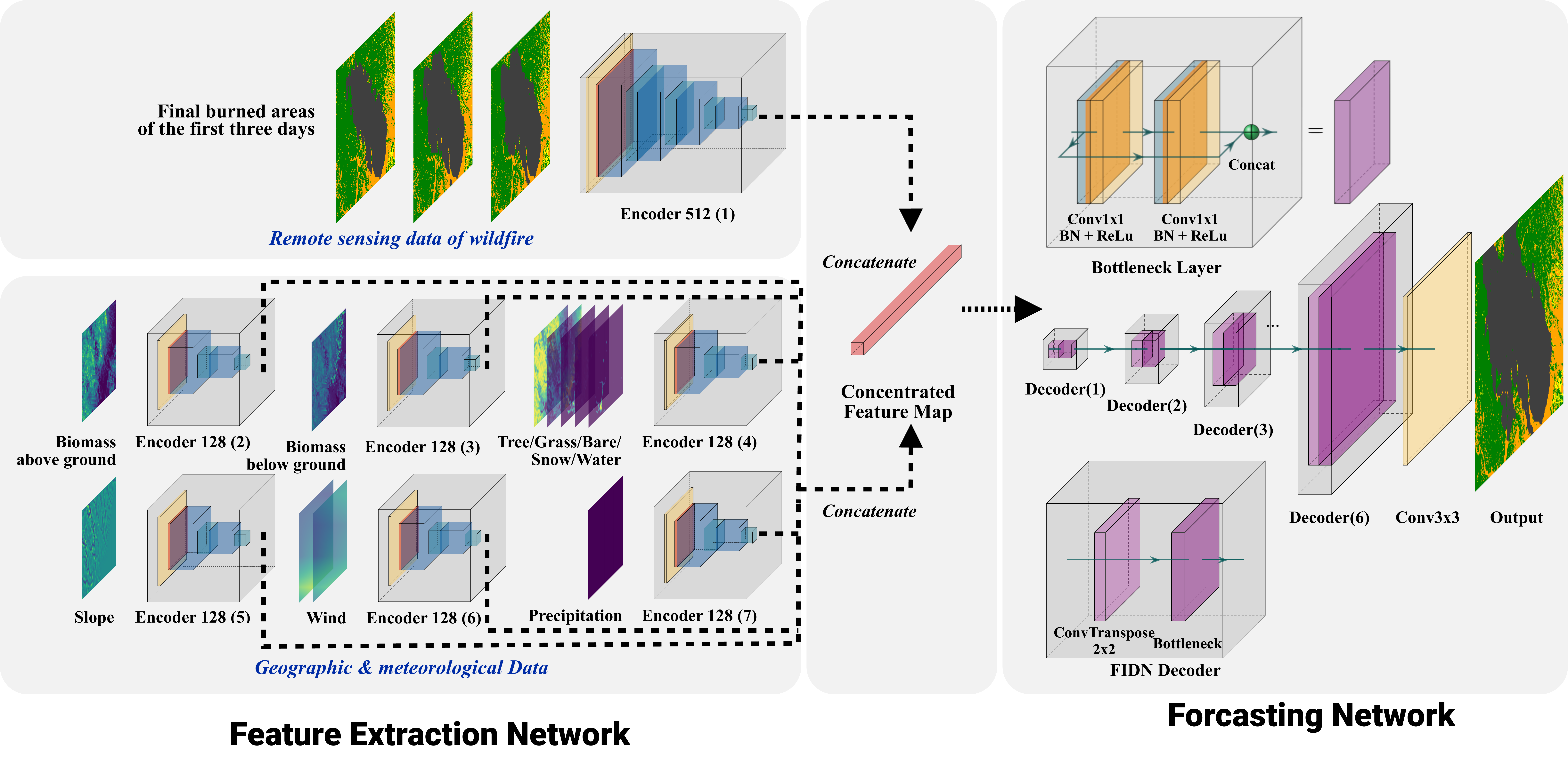}
    \end{center}
    \caption{\label{fig:model_structure} The structure of the Fire-Image-DenseNet (FIDN) model includes the feature extraction network and the forecasting network. }
\end{figure}

The Fire-Image-DenseNet (FIDN) model is designed for wildfire prediction and consists of two main components: a feature extraction network and a forecasting network. The FIDN model inputs consist of two types of data: \textit{Remote Sensing Data of Wildfires} (fireburnt areas of the first three days after ignition) and relevant \textit{Geographic and Meteorological Data} in the corresponding ecoregions including vegetation density, biomass carbon density, forest and grassland distribution, slope, wind angle and velocity, and precipitation. For an explanation of the rationale behind selecting these parameters, please refer to Section~\ref{section:2}. The overall structure of the FIDN model is depicted in Figure~\ref{fig:model_structure}.

\subsubsection{Feature Extraction Network: FIDN Encoder}

The feature extraction network is responsible for extracting and concatenating feature maps from various input images of all 15 layers, which are then fed into the forecasting network. The network is built using DenseNet architecture, specifically adapted for wildfire prediction by removing the top classification layer.

Each FIDN Encoder consists of convolutional layers followed by Dense blocks and Transition layers. To introduce non-linearity in the model and address the vanishing gradient problem, we apply the Rectified Linear Unit(ReLU) activation function after Batch Normalization in all convolution and dense block contexts throughout the model.

As mentioned above, The feature extraction network accommodates two types of inputs: Remote Sensing Data of Wildfires and Geographic and Meteorological Data. Based on the objectives of feature extraction, we have selected different input resolutions for these data types.

For remote sensing images of wildfire burnt areas, which contain critical details essential for capturing the dynamic changes of wildfires, we have chosen a relatively high resolution. This choice facilitates the preservation of spatial and visual details, thereby enabling more precise feature extraction. In this study, we use a resolution of 512x512 for this purpose.

Conversely, for images representing other vegetation, tropological and meteorological features, such as biomass, slope, and wind, we aim to reduce computational complexity during feature extraction while ensuring that these features are effectively processed and integrated. Therefore, we standardize these features to a smaller, uniform resolution of 128x128. By handling them separately, we ensure that each data type is appropriately processed, preserving the integrity of the information. This method allows each type of data to be processed and integrated effectively, considering their unique characteristics and resolutions.

To accommodate these different image dimensions and types, we employ two variants of the FIDN Encoder: \textit{FIDN Encoder-512} and \textit{FIDN Encoder-128}. It is important to note that while we have chosen input dimensions of 512x512 and 128x128 in this design, the architecture and methodology are highly versatile and flexible. Researchers can adjust input resolutions and encoder structures based on specific research requirements and data characteristics to suit different datasets and prediction tasks. This design approach is adaptable to images of other dimensions and can be expanded to develop additional variants. 

The following is an introduction to the two variants used in this research: 
\begin{figure}[!htb]
    \centering
    \begin{subfigure}[t]{1\textwidth}
        \centering
        \includegraphics[width=\linewidth]{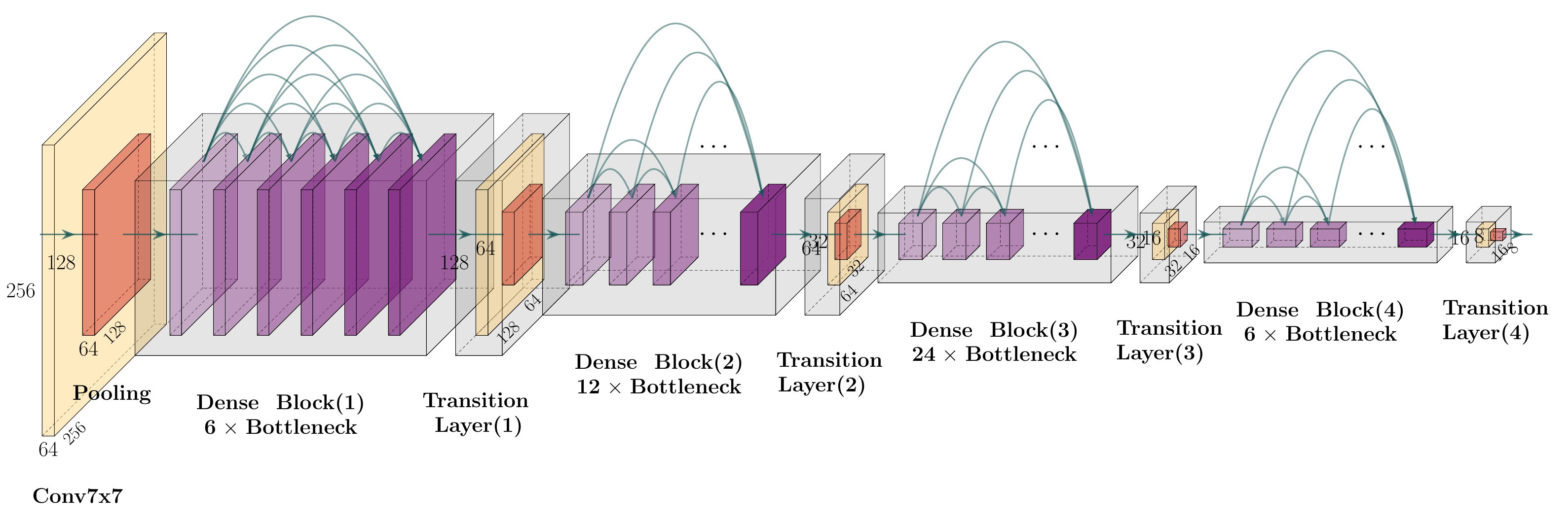}
        \caption{FIDN Encoder-512}
    \end{subfigure}%

    \begin{subfigure}[t]{0.6\textwidth}
        \centering
        \includegraphics[width=\linewidth]{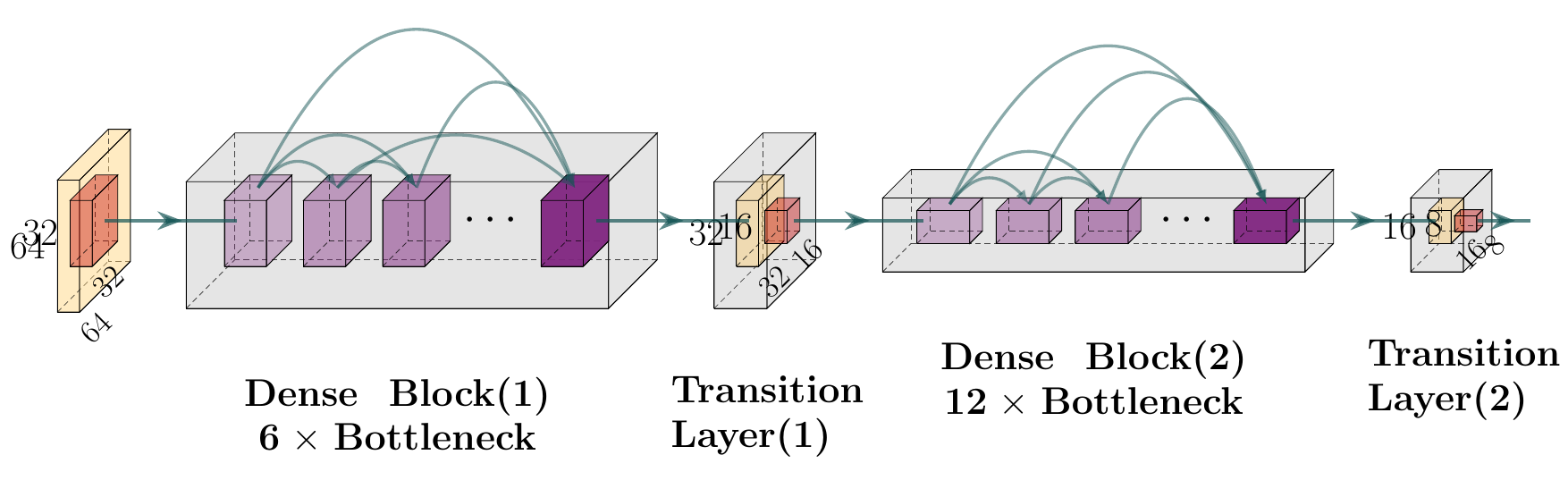}
        \caption{FIDN Encoder-128}
    \end{subfigure}%
    \caption{\label{fig:encoder} The details of FIDN: (a) FIDN encoder-128 and (b) FIDN encoder-512.}
\end{figure}

\textbf{FIDN Encoder-512} takes an image of dimension 512$\times$512 (i.e., input burnt area, see Table~\ref{table:datastructure}) as input, consisting of a convolution layer, a Polling layer, four sets of Dense blocks and Transition Layers. The four Dense blocks contain 6, 12, 24 and 6 Bottleneck layers respectively. The final output consists of feature maps with a dimension of 8$\times$8.

\textbf{FIDN Encoder-128} takes an image of dimension 128$\times$128 (i.e., vegetation, tropological and climate features, see Table~\ref{table:datastructure}) as input, consisting of a convolution layer, a Polling layer, two sets of Dense blocks and Transition Layers. The two Dense blocks contain 6 and 12 Bottleneck layers respectively. The final output is feature maps also with a dimension of 8$\times$8.

Finally, the feature extraction network, illustrated in Figure~\ref{fig:model_structure}, consists of:

\begin{itemize}
    \item one FIDN Encoder-512 sub-network for extracting features from $F^{(0)}$, $F^{(1)}$, $F^{(2)}$ and $F^{(n_k)}$
    \item six FIDN Encoder-128 sub-networks for extracting features from above ground biomass carbon density, below ground biomass carbon density, slope, tree/grass/smooth ground/snow/water, wind, and precipitation images respectively.
\end{itemize}

The encoded features are then concatenated and passed to the prediction network.

\subsubsection{Forecasting Network: FIDN Decoder}

The \textbf{FIDN Decoder} aims to predict the final burnt area by processing the concatenated features through a series of deconvolutional layers and Conv Blocks. The structure of the decoder includes: a deconvolution layers (Conv2DTranspose) with kernel sizes of 2 and strides of 2, used to upsample the feature maps gradually. The deconvolution layer is followed by a Conv Block, which includes a convolutional layer with a 3 $\times$ 3 window and ReLU activation. for data ascension and valid information separation.

The encoded input (after the feature extraction network) is passed through 5 FIDN Decoders sequentially. The final output layer applies a sigmoid activation function to produce a predicted image with dimensions 512 $\times$ 512.

\subsection{Loss Function and Metrics}
To accurately assess our model's performance in predicting binary images of wildfire burnt areas, we employ a combination of evaluation metrics: Binary Cross-Entropy (BCE), Mean Squared Error (MSE), Root Relative Squared Error (RRMSE), Structural Similarity Index Measure (SSIM), and Peak Signal-to-Noise Ratio (PSNR). The use of binary representation frames our task as a binary classification problem for each pixel, making BCE and MSE essential for evaluating classification accuracy and prediction error. Additionally, RRMSE, SSIM, and PSNR are chosen to measure the visual quality and structural integrity of the predicted images against actual satellite observations.

For the sake of notation, in the following equations, we assume a preprocessed wildfire burnt area image consists of $N \times M$ pixels. $(i, j)$ represents the pixel coordinates in the image with $0 \leq i \leq N, 0 \leq j \leq M$. $\mathcal{F}^{(t)}_{k}$ denotes the true burnt image observed by the satellite on day $t$. $\mathcal{F}^{(n_k)}_{k_{ij}}$ is a binary number representing the burn information at pixel $(i, j)$ on day $n_k$. The predicted burnt status (i.e., the output of predictive models) is denoted by $\hat{\mathcal{F}}^{(n_k)}_{k_{ij}}$.

1. \textbf{Binary Cross-entropy(BCE)}
Binary Cross-entropy is a loss function commonly used in binary classification problems~\citep{ho2019real}. The formula for this algorithm is shown in Eq.~\ref{eq:bce}.
\begin{equation}
    BCE = - \frac{1}{N \times M}\sum_{i=0}^{N}\sum_{j=0}^{M}\mathcal{F}^{(n_k)}_{k_{ij}} \cdot \log(\hat{\mathcal{F}}^{(n_k)}_{k_{ij}}) + (1 - \mathcal{F}^{(n_k)}_{k_{ij}}) \cdot log(1-\hat{\mathcal{F}}^{(n_k)}_{k_{ij}}),
    \label{eq:bce}
\end{equation}

When Binary Cross-entropy is selected as the loss function, this prediction task can be regarded as a binary classification problem on the pixel level, predicting whether the region represented by each pixel has been burnt out or not.

2. \textbf{Mean Squared Error(MSE)}
Mean Squared Error measures the average squared difference between the estimated and actual values, a long-established metric for evaluating the similarity of images~\citep{marmolin1986subj} and physical fields~\cite{xu2024comprehensive, fu2023hierarchical}. The MSE formula is presented in Eq.~\ref{eq:mse}.
\begin{equation}
    MSE = \frac{1}{N \times M}\sum_{i=0}^{N}\sum_{j=0}^{M}(\mathcal{F}^{(n_k)}_{k_{ij}}-\hat{\mathcal{F}}^{(n_k)}_{k_{ij}})^2
    \label{eq:mse}
\end{equation}

3. \textbf{Relative Root Mean Squared Error(RRMSE)}
Root Mean Squared Error (RMSE) is the square root of MSE, while Relative Root Mean Squared Error (RRMSE) is the dimensionless form of RMSE as formulated using Eq.~\ref{eq:rrmse}.
\begin{equation}
    RRMSE = \sqrt{\frac{MSE}{\sum_{i=0}^{N}\sum_{j=0}^{M}(\hat{\mathcal{F}}^{(n_k)}_{k_{ij}})^2}}
    \label{eq:rrmse}
\end{equation}

4. \textbf{Structural Similarity(SSIM)}
Structural Similarity is a measure of the similarity between two images~\citep{wang2004image}. For simplicity, here we denote $x$ as the actual field of burnt status $\mathcal{F}^{(t)}_{k_{ij}}$ and $y$ as the predicted burnt status $\hat{\mathcal{F}}^{(t)}_{k_{ij}}$, the SSIM between them can then be formalised by Eq.~\ref{eq:ssim}.
\begin{equation}
    SSIM(x,y)=\frac{(2\mu_x+\mu_y+c_1)(2\sigma_{xy}+c_2)}{(\mu_x^2 + \mu_y^2+c_1)(\sigma_x^2+\sigma_y^2+c_2)},
    \label{eq:ssim}
\end{equation}
where $(\mu_x, \sigma_x^2)/(\mu_y, \sigma_y^2)$ denote the mean and the variance of $x$ and $y$ respectively, $\sigma_{xy}$ denotes the covariance of $x$ and $y$, $c_1$ and $c_2$ are constant coefficients for the positional stability. According to~\citep{sara2019image}, SSIM is capable of giving perception-based errors whereas MSE only estimates absolute errors.

5. \textbf{Peak signal-to-noise ratio (PSNR)}
Peak signal-to-noise ratio, as defined in Eq.~\ref{eq:psnr}, is also a well-known metric for image similarity~\citep{hore2010image}.
\begin{equation}
\begin{aligned}
    PSNR &= 10 \cdot \log_{10} \left( \frac{\mathit{Max}(\mathcal{F}^{(t)}_k)^2}{\mathit{MSE}} \right)
    \label{eq:psnr}
\end{aligned}
\end{equation}
where $\mathit{Max}(\mathcal{F}^{(t)}_k)$ is the maximum value in the final burnt area. Compared to MSE~\citep{sara2019image}, PSNR is also capable of giving perception-based errors.

\section{Study area and Data curation}
\label{section:2}
In this section, we describe the data sources and preprocessing methods used to train and test FIDN.

Formally, for a wildfire event indexed $k$ of duration $n_k$ days, $\{\mathcal{F}^{(t)}_{k}\}_{t=1,...,n_k}$ denotes the burnt area on day $t$, which is defined on a two-dimensional grid. $\mathcal{F}^{(t)}_{k} \in \mathbb{R}^{N_k} \times \mathbb{R}^{M_k}$ where $N_k \times M_k$ is the dimension of the ecoregion. Each point in the grid $\mathcal{F}^{(t)}_{k_{ij}} \quad (0 \leq i \leq N_k, 0 \leq j \leq M_k)$ is represented in binary numbers, where 0 for not burnt and 1 for burnt. This approach streamlines the model's focus on predicting the final burnt area, simplifying input complexity and enhancing both training and prediction efficiency. Such binary simplification reduces the computational load, crucial for accurately forecasting fire spread with clear target states. Additionally, acquiring and processing more detailed wildfire-related parameters is notably time-consuming, making the binary representation advantageous by mitigating the extensive time and resources required for data collection and preparation.

The FIDN model takes the burnt area of the first three days after ignition (i.e. $\mathcal{F}^{(0)}_{k}$, $\mathcal{F}^{(1)}_{k}$ and $\mathcal{F}^{(2)}_{k}$) as input and outputs the final burnt area ($\mathcal{F}^{(n_k)}_{k}$). The data of $\{\mathcal{F}^{(t)}_{k}\}_{t=1,...,n_k}$ is extracted from the daily fire perimeter database generated from the Moderate Resolution Imaging Spectroradiometer (MODIS) and the Visible Infrared Imaging Radiometer Suite (VIIRS) active fire products~\cite{scaduto2020satellite}. 
VIIRS detected hot spots twice a day on a global scale at a resolution of $275m$ while MODIS provides hot spot detection 4 times a day globally ~\citep{giglio2016collection}. To characterize fire spread at a daily time scale, the natural neighbor geospatial interpolation method was used to interpolate the discrete active fire points detected by these two sensors, and the interpolated geometries were further simplified to polygons using the polynomial approximation with exponential kernel technique,  following the method developed and validated by Scaduto et al. (2019).  Since we focus on large wildfire events, we selected the fire events with $n_k > 4$. We extracted a total of 333 fire events that occured in western US from 2017 to 2019. 
To maintain the relative size of the wildfires geographically, the images of the burnt area are reshaped using the same scale roughly 0.026 $km^2$ per pixel according to latitude and longitude distance, resulting in two-dimensional vectors of size 512 $\times$ 512 for all fire events. The final area burned was used as the response variable, while the sequence of the areas burned during the first three days of fire events were used as one of the inut datasets.  

\begin{figure}[!htb]
    \begin{center}
        \includegraphics[width=1\textwidth]{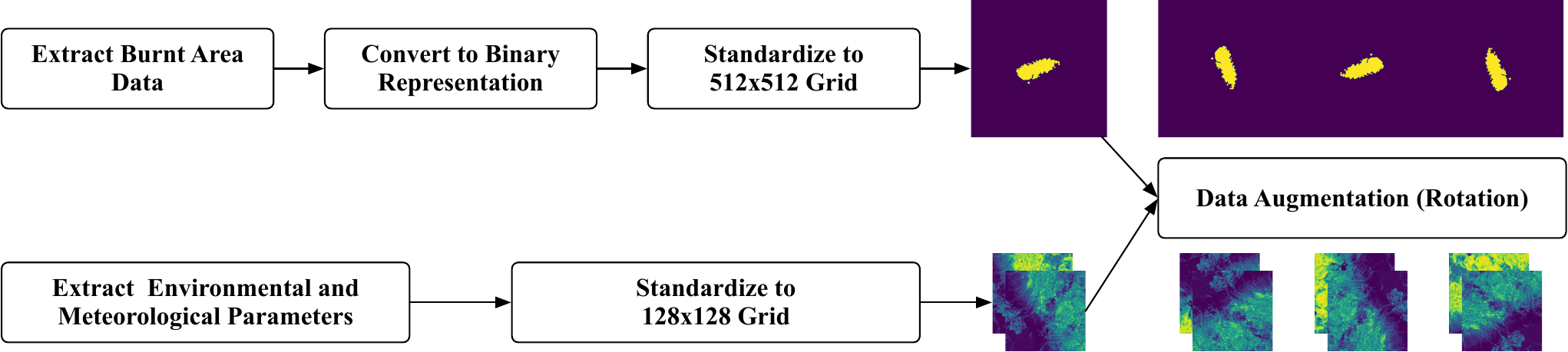}
    \end{center}
    \caption{Data Preprocessing Workflow for Wildfire Prediction Model}
    \label{fig:preprocessing}
\end{figure}

We also extracted environmental and meteorological parameters that have been shown in other studies to have a strong influence on fire occurrence and spread~\citep{alexandridis2008cellular,schroeder2014new,just2016fire,trucchia2020propagator}, including vegetation density, biomass carbon density, forest and grassland distribution, slope, wind angle and velocity, and precipitation. We also included information on the distribution of non-flammable materials such as snow, water, and bare ground. These parameters were extracted from several satellites and observation sources (Table~\ref{table:datastructure}). All input data were pre-processed to a common 128 $\times$ 128 grid. The flow of data preprocessing is shown in Figure \ref{fig:preprocessing}.

\begin{figure}[!htb]
    \begin{center}
        \includegraphics[width=1\textwidth]{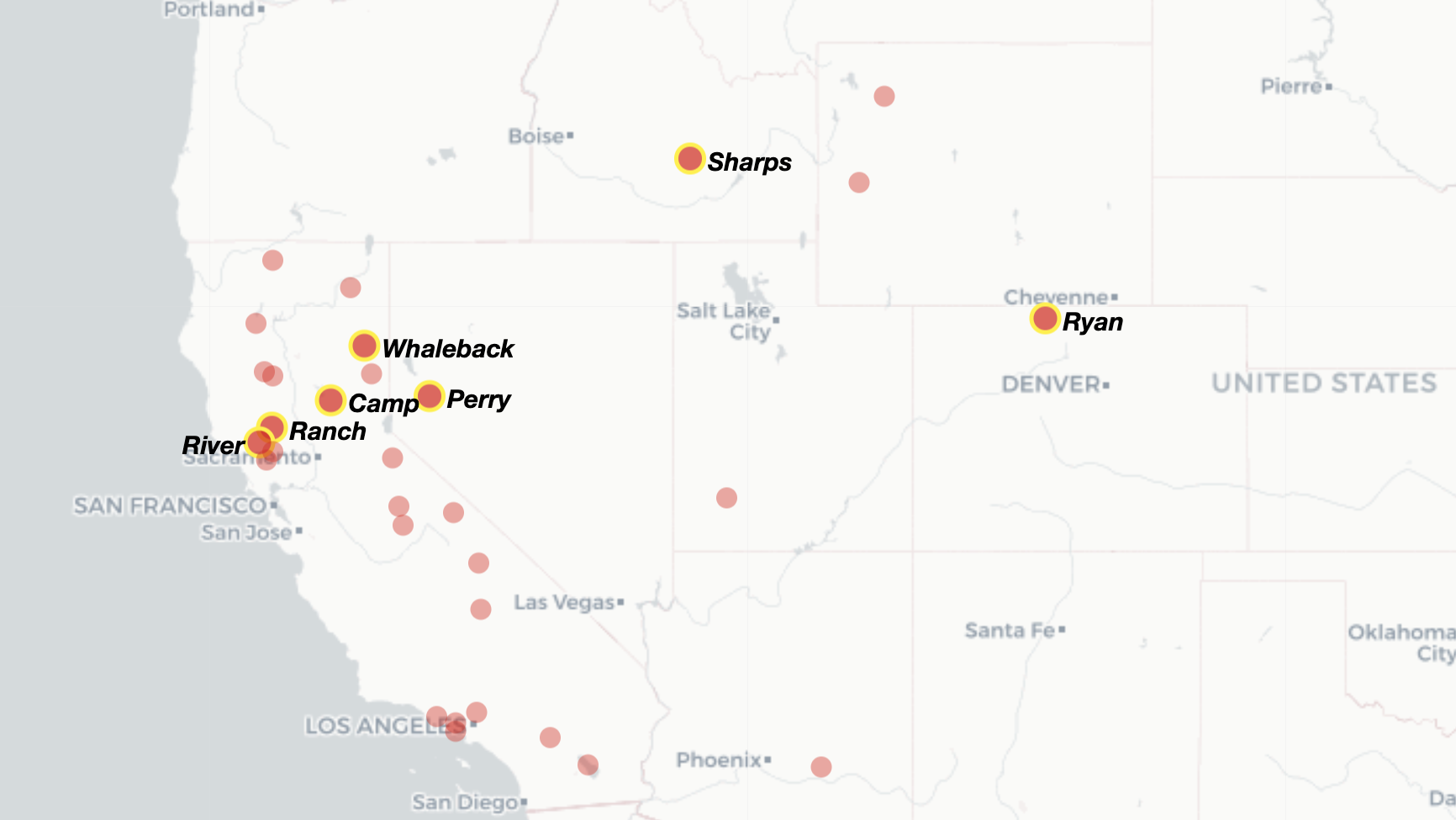}
    \end{center}
    \caption{The location of the wildfires in the test dataset. Each fire is shown as a coloured dot and the seven fires used in the detailed analysis of the impact of size and duration on model prediction are named.
    }
    \label{fig:location}
\end{figure}

The FIDN model is trained and tested using chronological historical wildfire data from the western US (mainly California). We construct training and testing datasets by extracting 303 fire events that occurred in the western US with a final burnt area $>100km^2$. The training dataset consists of 243 wildfire events that occurred between 2012 and 2017. The validation and test datasets each consist of 30 wildfire events from the following years, i.e., 2018 to 2019. Further information on the location and characteristics of the wildfires in the test data is shown in Figure~\ref{fig:location}. Due to the relatively small total number of wildfire events in the dataset, only 243 fire events are available to train FIDN. Data augmentation is performed by rotating the two-dimensional fields in the training set by 90, 180 and 270 degrees. The geographical and meteorological data have been rotated accordingly. The final training set contains a total of 972 augmented fire events. Table \ref{tab:wildfire_data} provides detailed information regarding the sample datasets.

\begin{table}[h]
    \centering
    \resizebox{\textwidth}{!}{%
    \begin{tabular}{cccccc}
        \toprule
        \textbf{Dataset Name}       & \textbf{Number of Events} & \textbf{Training Set Size} & \textbf{Augmented Training Set Size} & \textbf{Validation Set Size} & \textbf{Test Set Size} \\
         \midrule
        Wildfire Events    & 303              & 243                & 972 & 30                  & 30            \\
        \bottomrule
    \end{tabular}
    }
    \caption{Detailed Information of Sample Datasets}
    \label{tab:wildfire_data}
\end{table}

\begin{table}[!hbtp]
    \raggedleft
    \resizebox{\textwidth}{!}{%
    \begin{tabular}{lllll}
        \toprule
        \textbf{Channel No.} & \textbf{Description} & \textbf{Source} & \textbf{Resolution} & \textbf{Dimension} \\
        \midrule
        1 & $\mathcal{F}^{(0)}_k$ (burnt area in day 0) & MODIS~\citep{giglio2016collection}/VIIRS & $\thickapprox 275m$ & 512 $\times$ 512 \\
        2 & $\mathcal{F}^{(1)}_k$ (burnt area in day 1) & MODIS/VIIRS & $\thickapprox 275m$  & 512 $\times$ 512  \\
        3 & $\mathcal{F}^{(2)}_k$ (burnt area in day 2)& MODIS/VIIRS & $\thickapprox 275m$  & 512 $\times$ 512  \\
        4 & biomass above ground & ORNL  DACC~\citep{spawn2020global} & 300 m & 128 $\times$ 128  \\
        5 & biomass below ground & ORNL  DACC & 300 m & 128 $\times$ 128 \\
        6 & slope & CSP~\citep{theobald2015ecologically} & 270 m & 128 $\times$ 128 \\
        7 & tree density & PROBA-V~\citep{buchhorn2020copernicus} & 100 m & 128 $\times$ 128 \\
        8 & grass density & PROBA-V & 100 m & 128 $\times$ 128 \\
        9 & bare density & PROBA-V & 100 m & 128 $\times$ 128 \\
        10 & snow density & PROBA-V & 100 m & 128 $\times$ 128 \\
        11 & water density & PROBA-V & 100 m & 128 $\times$ 128 \\
        12 & 10m u-component of wind(monthly average) & ERA5~\citep{hersbach2018era5} & 27830 m & 128 $\times$ 128 \\
        13 & 10m v-component of wind(monthly average) & ERA5~\citep{hersbach2018era5} & 27830 m & 128 $\times$ 128 \\
        14 & total precipitation(rainfall + snowfall) (monthly sums) & ERA5~\citep{hersbach2018era5} & 27830 m & 128 $\times$ 128\\
        15 & $\mathcal{F}^{(n_k)}$(final burnt area)& MODIS & $\thickapprox 275m$ & 512 $\times$ 512  \\
        \bottomrule
    \end{tabular}
    }
    \caption{List of datasets used in the study, including information on the source and resolution. The dimensionality is given after the dataset has been pre-processed.}
    \label{table:datastructure}
\end{table}

\section{Numerical Results}
\label{section:4}
In this section, we present and analyze the FIDN model proposed in this paper for predicting the final burnt area of wildfires in the test dataset. The performance of the proposed approach is compared against the state-of-the-art CA~\citep{alexandridis2008cellular} and MTT~\citep{finney2002fire} models.

As mentioned in Section 2, We train our predictive model, FIDN, using the daily burnt area for 243 individual wildfires from the western US that occurred between 2012 and 2017. The corresponding environmental and climate variables are considered as model inputs. We evaluate the performance of the model using a validation dataset of 30 wildfires from the same period and an independent test dataset of 30 wildfires that occurred in 2018 and 2019.

\begin{figure}[!htb]
    \centering
    \begin{subfigure}[t]{0.43\textwidth}
        \centering
        \includegraphics[width=\linewidth]{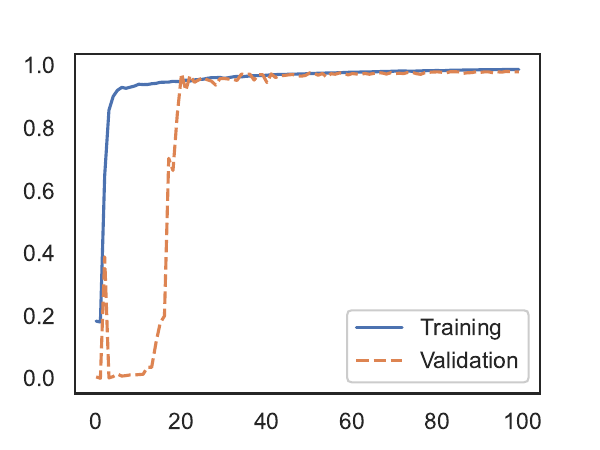}
        \caption{SSIM}
    \end{subfigure}%
    \begin{subfigure}[t]{0.43\textwidth}
        \centering
        \includegraphics[width=\linewidth]{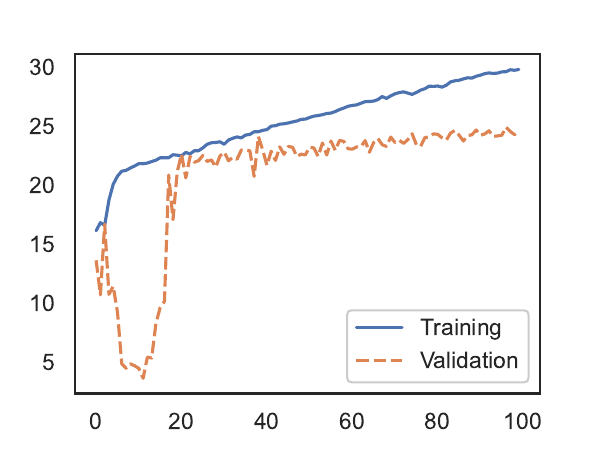}
        \caption{PSNR}
    \end{subfigure}%

    \begin{subfigure}[t]{0.43\textwidth}
        \centering
        \includegraphics[width=\linewidth]{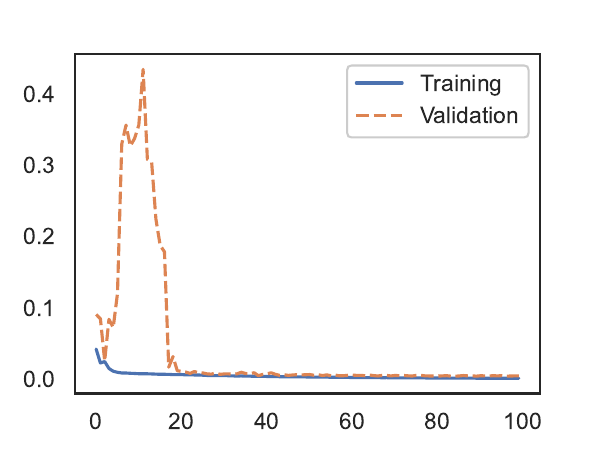}
        \caption{MSE}
    \end{subfigure}%
    \begin{subfigure}[t]{0.43\textwidth}
        \centering
        \includegraphics[width=\linewidth]{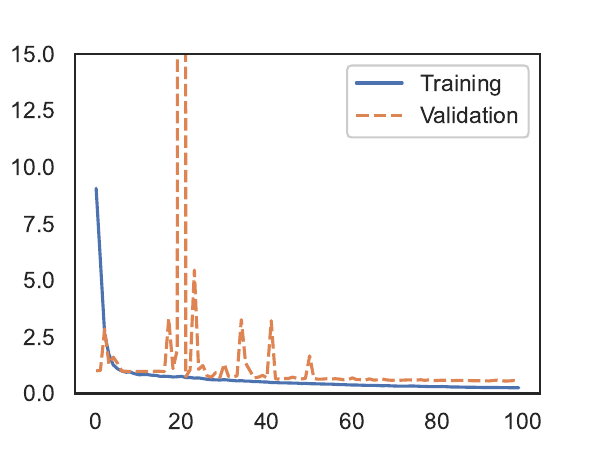}
        \caption{RRMSE}
    \end{subfigure}%

    \begin{subfigure}[t]{\textwidth}
        \centering
        \includegraphics[width=\linewidth]{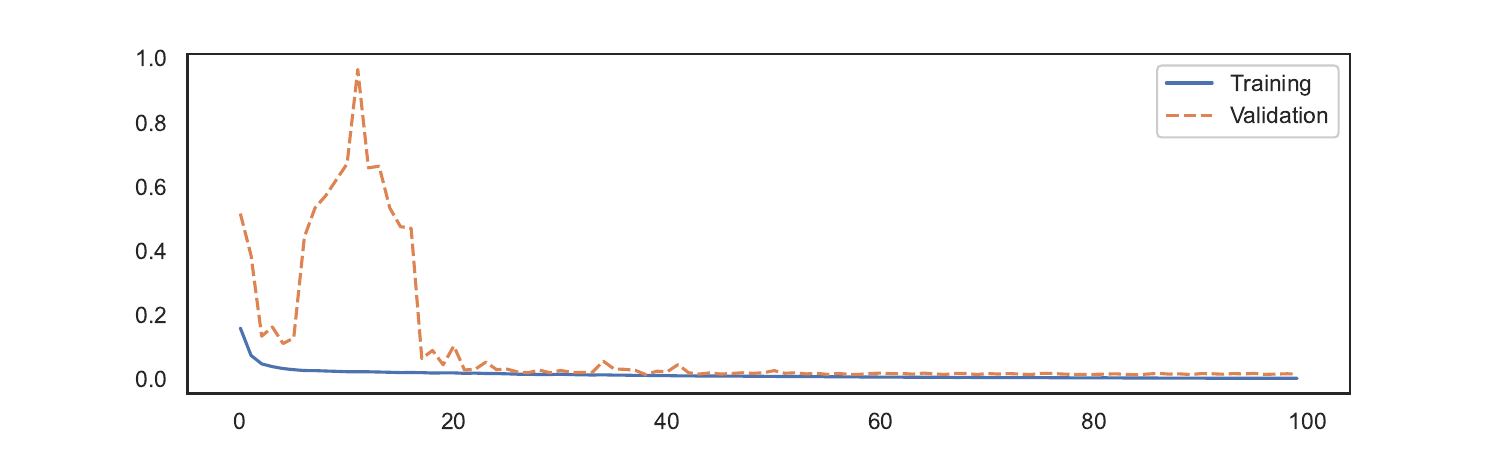}
        \caption{Training Loss (BCE)}
    \end{subfigure}%
    \caption{Metrics of model performance during training and validation against the number of training epochs. The five metrics are (a) the structural similarity index measure (SSIM), which is a normalised value between 1 for perfect correlation and 0 for no correlation, (b) the peak signal to noise ratio (PSNR) in $km^2$, (c) the mean square error (MSE) in $km^2$, (d) the relative root mean square error (RRMSE) in $km^2$, and (e) training loss (Binary Cross Entropy, BCE loss) where a perfect model has a BCE of 0.}
    \label{fig:mertic_change}
\end{figure}

Figure~\ref{fig:mertic_change} shows the evolution of BCE loss and other metrics during the training process. There is a steady improvement in model performance within each training epoch. On the validation set, there is a sharp protrusion in the first 20 epochs, after which the loss starts to drop steeply and finally stabilizes. After 40 epochs, while the metrics measured on the training set continue to progress, those metrics evaluated on the validation set remain stable and improve slightly. Overall we observe that as the metrics of the training set improve, the results of the validation set remain stable and increasing (i.e. no over-fitting occurred). As explained in the work of~\citep{huang2017densely}, dense connections have a regularization effect and can reduce the model over-fitting. The latter is extremely important for this study since the size of the data set is small, leading to a high risk of training overfitting~\citep{ying2019overview}.

We compare the performance of the proposed FIDN model against  CA and FlamMap MTT models. All experiments were conducted on the same computer system to ensure consistency and reliability in the performance evaluation. The system specifications included an Intel i9-13900KF processor, 64GB of DDR5 RAM, and an Nvidia RTX 4090 graphics card with 24GB of VRAM, providing a fair and unbiased comparison of the FIDN, CA, and FlamMap MTT models. For a fair comparison, all simulations started from day 2 after ignition, $\mathcal{F}^{(2)}_k$, since this information is given to the FIDN model. Both CA and MTT models used the same environmental data (tree, slope, density) as FIDN, ensuring consistency. For wind data, $u$ and $v$ vector components were calculated from two-dimensional data provided to FIDN and directly used as input for the CA model. For the MTT model, wind speed and direction were further computed from these $u$ and $v$ components. MTT also applied the "Finney" method for Crown Fire Calculation \citep{finney2003calculation}, with other settings at defaults.  Furthermore, precise wildfire durations are provided specifically for CA and MTT models, but this is unrealistic in actual fire events.

Table~\ref{table:avg_mertic} shows that the FIDN model produces more accurate predictions of the final burnt area compared to CA and MTT on all metrics. The MSE for FIDN is about 82\% lower than the CA model and 67\% lower than the FlamMap MTT model. At the same time, the SSIM of FIDN has an average of 97\% (with a very small standard deviation) which is 6\% higher than the CA model and 2\% higher than the FlamMAP MTT model. This improved performance is accompanied by a significant reduction in computational time. FIDN is about three orders of magnitude faster than either the CA or MTT. Thus, FIDN is capable of providing predictions closer to the observed burnt area, with a considerably lower computational cost.

\begin{table}[]
    \centering
    \resizebox{\textwidth}{!}{
    \begin{tabular}{@{}llllll@{}}
    \toprule
     &  & \textbf{FIDN} & \textbf{CA from $\mathcal{F}^{(2)}$} & \textbf{CA from $\mathcal{F}^{(0)}$} & \textbf{MTT} \\ \midrule
    \multirow{2}{*}{\textbf{SSIM}} & Mean ± stddev & \textbf{0.971 ± 0.015} & 0.910 ± 0.090 & 0.912 ± 0.090 & 0.953 ± 0.050 \\
     & Median (IQR) & 0.974 (0.015) & 0.948(0.144) & 0.947(0.134) & \textbf{0.978 (0.068)} \\ \midrule
    \multirow{2}{*}{\textbf{PSNR}} & Mean ± stddev & \textbf{20.993 ± 2.787} & 17.029 ± 8.127 & 16.623 ± 7.411 & 19.792 ± 7.617 \\
     & Median (IQR) & \textbf{21.473 (3.638)} & 16.105   (14.067) & 14.923   (12.110) & 19.640   (13.172) \\ \midrule
    \multirow{2}{*}{\textbf{MSE}} & Mean ± stddev & \textbf{0.010 ± 0.008} & 0.056 ± 0.063 & 0.056 ± 0.064 & 0.031 ± 0.037 \\
     & Median (IQR) & \textbf{0.007   (0.006)} & 0.025   (0.099) & 0.032   (0.085) & 0.011   (0.046) \\ \midrule
    \multirow{2}{*}{\textbf{RRMSE}} & Mean ± stddev & \textbf{0.825 ± 0.212} & 0.897 ± 0.527 & 2.168 ± 4.735 & 1.017 ± 0.666 \\
     & Median (IQR) & 0.821   (0.267) & \textbf{0.797   (0.155)} & 0.842 (0.233) & 0.867   (0.201) \\ \midrule
    \multirow{2}{*}{\textbf{Time}} & Mean ± stddev & \textbf{1.127 ± 0.026} & 1577.614 ± 3399.791 & 1901.027 ± 3807.586 & 3419.216 ± 5532.844 \\
     & Median (IQR) & \textbf{1.125 (0.032)} & 369.345 (1320.606) & 486.698 (1539.896) & 656.345 (3952.790) \\ \bottomrule
    \end{tabular}
    }
    \caption{Performance statistics (mean, median and standard deviation) of model predictions of final burnt area summarized over 23 wildfires from the test dataset. The performance of the Fire-Image-DenseNet (FIDN) model is compared to predictions of the cellular automaton (CA) model and the FlamMap Minimum Travel Time (MTT) model. The five evaluation metrics are included: the structural similarity index measure (SSIM), a normalised value between 1 for perfect correlation and 0 for no correlation; the peak signal to noise ratio (PSNR); the mean square error (MSE) in $km^2$ ; the relative root mean square error (RRMSE) and the online runtime for burnt area prediction.}
    \label{table:avg_mertic}
\end{table}

In addition, we further analyze the model outputs with geographical information (such as forest and grassland in the corresponding ecoregion) to interpret the models' decision strategies and examine their ability to handle fuel information (combustible/non-combustible). We have selected seven representative fire events in the test set, based on their final burnt area size and duration, as shown in Table~\ref{table:special_fire}. The different metrics regarding the three models' prediction performance are presented in Table~\ref{table:special_fire_mertic}. The fire events in Table~\ref{table:special_fire} and Table~\ref{table:special_fire_mertic} are divided into three categories (i.e., large, moderate and small fires) based on the observed final burnt area.

\begin{table}[!htb]
    \centering
    \resizebox{\textwidth}{!}{%
    \begin{tabular}{lccccrr}
    \toprule
    \textbf{Type} & \textbf{Fire name} & \textbf{Year} & \textbf{Longitude} & \textbf{Latitude} & \textbf{Area($km^2$)} & \textbf{Duration (day)} \\
    \midrule
    \multirow{2}{*}{\textbf{Large}} & Camp      & 2019 & -121.56   & 39.75    & 1178.53                     & 13             \\
    & Ranch     & 2018 & -122.78   & 39.29    & 3064.08                     & 48             \\ \midrule
    \multirow{3}{*}{\textbf{Moderate}} & Perry     & 2018 & -119.49   & 39.80     & 453.12                      & 6              \\
    & Sharps    & 2018 & -114.05   & 43.53    & 437.61                      & 9              \\
    & River     & 2018 & -123.03   & 39.05    & 404.53                      & 11             \\ \midrule
    \multirow{2}{*}{\textbf{Small}} & Whaleback & 2018 & -120.83   & 40.63    & 116.84                      & 6              \\
    & Ryan      & 2018 & -106.61   & 41.03    & 299.14                      & 16            \\
    \bottomrule
    \end{tabular}
    }
    \caption{ Details of the seven fires from the test dataset used for analysis of the impact of fire size and duration on model performance.}
    \label{table:special_fire}
\end{table}

\begin{figure}[!htbp]
    \centering
    \begin{subfigure}[t]{0.5\textwidth}
        \centering
        \includegraphics[width=\linewidth]{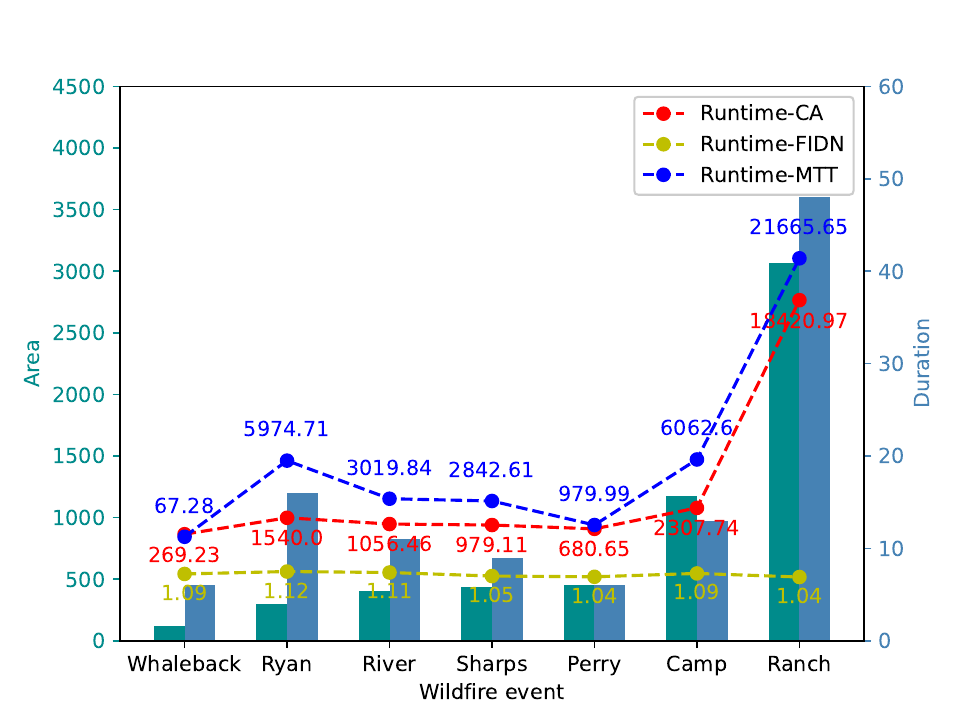}
        \caption{Runtime}
    \end{subfigure}%
    \begin{subfigure}[t]{0.5\textwidth}
        \centering
        \includegraphics[width=\linewidth]{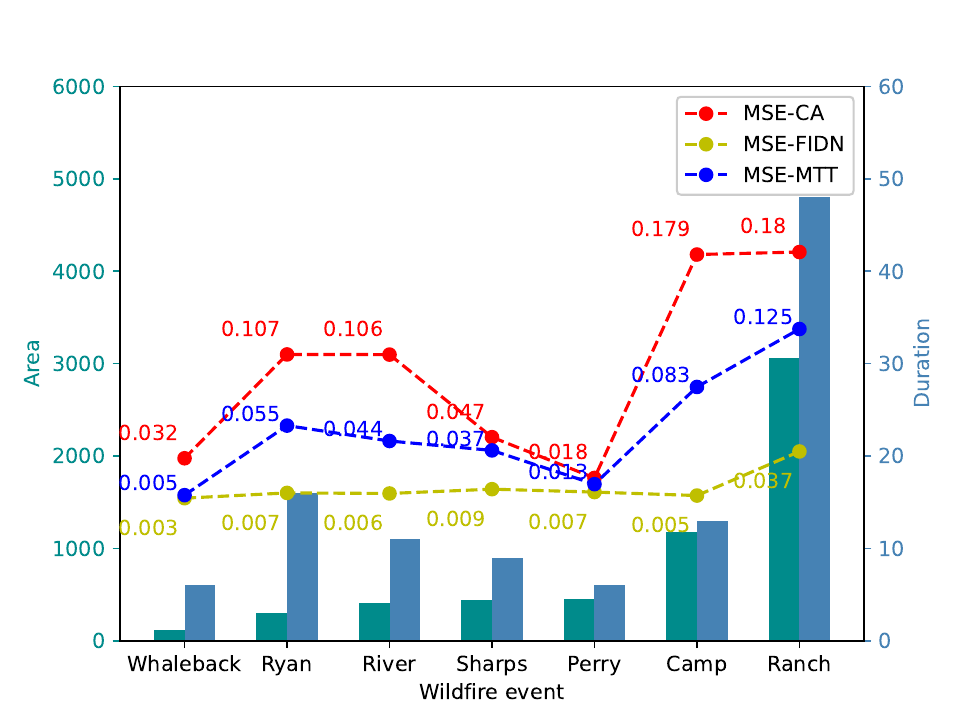}
        \caption{MSE}
    \end{subfigure}%
    \caption{Model performance, as measured by (a) the mean square error (MSE) in $km^2$ and (b) computation time in seconds, with respect to fire size and fire duration, for seven wildfires from the test dataset, for three models: the Fire-Image-DenseNet (FIDN) model, the cellular automaton (CA) model and the FlamMap Minimum Travel Time (MTT) model.}
    \label{fig:comp}
\end{figure}

\begin{table}[!htb]
    \centering
    \resizebox{\textwidth}{!}{%
    \begin{tabular}{@{}cc|rr|rrr|rr@{}}
    \toprule
    \multicolumn{1}{l}{\textit{\textbf{}}} &
      \textit{\textbf{}} &
      \multicolumn{2}{c|}{\textit{\textbf{Large}}} &
      \multicolumn{3}{c|}{\textit{\textbf{Moderate}}} &
      \multicolumn{2}{c}{\textit{\textbf{Small}}} \\ \cmidrule(l){3-9}
    \multicolumn{1}{l}{\textbf{}} &
      \textit{\textbf{}} &
      \multicolumn{1}{c}{\textbf{Camp}} &
      \multicolumn{1}{c}{\textbf{Ranch}} &
      \multicolumn{1}{c}{\textbf{Perry}} &
      \multicolumn{1}{c}{\textbf{Sharps}} &
      \multicolumn{1}{c|}{\textbf{River}} &
      \multicolumn{1}{c}{\textbf{Whaleback}} &
      \multicolumn{1}{c}{\textbf{Ryan}} \\ \midrule
    \multirow{3}{*}{\textbf{SSIM}} &
      \textit{FIDN} &
      \textbf{0.9699} &
      \textbf{0.9202} &
      \textbf{0.9711} &
      \textbf{0.9701} &
      \textbf{0.9722} &
      0.9840 &
      \textbf{0.9712} \\
     &
      \textit{CA} &
      0.7564 &
      0.6989 &
      0.9461 &
      0.9011 &
      0.8307 &
      0.9476 &
      0.8507 \\
     &
      \textit{MTT} &
      0.8718 &
      0.8291 &
      0.9691 &
      0.9409 &
      0.9289 &
      \textbf{0.9868} &
      0.9236 \\ \midrule
    \multirow{3}{*}{\textbf{PSNR}} &
      \textit{FIDN} &
      \textbf{23.1578} &
      \textbf{14.3639} &
      \textbf{21.3467} &
      \textbf{20.2535} &
      \textbf{21.9894} &
      \textbf{25.4088} &
      \textbf{21.7415} \\
     &
      \textit{CA} &
      7.4789 &
      7.4340 &
      17.5654 &
      13.2846 &
      9.7266 &
      14.9878 &
      9.7231 \\
     &
      \textit{MTT} &
      10.8014 &
      9.0324 &
      18.8604 &
      14.2758 &
      13.5509 &
      22.9404 &
      12.5756 \\ \midrule
    \multirow{3}{*}{\textbf{MSE}} &
      \textit{FIDN} &
      \textbf{0.0048} &
      \textbf{0.0366} &
      \textbf{0.0073} &
      \textbf{0.0094} &
      \textbf{0.0063} &
      \textbf{0.0029} &
      \textbf{0.0067} \\
     &
      \textit{CA} &
      0.1787 &
      0.1805 &
      0.0175 &
      0.0469 &
      0.1065 &
      0.0317 &
      0.1066 \\
     &
      \textit{MTT} &
      0.0831 &
      0.1250 &
      0.0130 &
      0.0374 &
      0.0441 &
      0.0051 &
      0.0553 \\ \midrule
    \multirow{3}{*}{\textbf{RRMSE}} &
      \textit{FIDN} &
      \textbf{0.3631} &
      1.1633 &
      0.6013 &
      \textbf{0.6336} &
      \textbf{0.5679} &
      \textbf{0.4968} &
      \textbf{0.7431} \\
     &
      \textit{CA} &
      0.7954 &
      \textbf{0.7975} &
      \textbf{0.4944} &
      0.6610 &
      0.8140 &
      0.7215 &
      0.8901 \\
     &
      \textit{MTT} &
      0.8206 &
      0.8138 &
      0.6162 &
      0.7564 &
      0.8197 &
      0.7665 &
      0.8967 \\ \midrule
    \multirow{3}{*}{\textbf{Runtime(s)}} &
      \textit{FIDN} &
      \textbf{4.13} &
      \textbf{1.04} &
      \textbf{1.04} &
      \textbf{1.05} &
      \textbf{1.11} &
      \textbf{1.09} &
      \textbf{1.12} \\
     &
      \textit{CA} &
      2307.74 &
      18420.97 &
      680.65 &
      979.11 &
      1056.46 &
      269.23 &
      1540.00 \\
     &
      \textit{MTT} &
      6062.60 &
      21665.65 &
      979.99 &
      2842.61 &
      3019.84 &
      67.28 &
      5974.71 \\ \bottomrule
    \end{tabular}%
    }
    \caption{Metrics for model performance against the seven fires from the test dataset used for the analysis of the impact of fire size and duration on model performance. The five metrics are (a) the structural similarity index measure (SSIM), which is a normalised value between 1 for perfect correlation and 0 for no correlation, (b) the peak signal to noise ratio (PSNR), (c) the mean square error (MSE) in $km^2$, (d) the relative root mean square error (RRMSE), and (e) the online runtime in seconds for fire prediction.}
    \label{table:special_fire_mertic}
\end{table}

As shown in Table~\ref{table:special_fire} and~\ref{table:special_fire_mertic}, both the computational efficiency and accuracy of CA and MTT models decrease significantly as the duration of the fire events increases. On the other hand, FIDN has a more consistent performance, as depicted in Figure~\ref{fig:comp} where the fire events are listed in the increasing order of final burnt area size. For instance, both the Whaleback and Perry fire events had a duration of six days, but the Perry fire, being four times larger in terms of the final burnt area compared to the Whaleback fire, led to a significant increase in the execution time required for CA and MTT simulations. For significantly larger fire events, such as Ranch, the simulation time for CA and MTT exceeds 5 hours. On the other hand, the FIDN model significantly reduces the online computational time to about 1 second for all the fire events. Moreover, CA and MTT models face significant challenges in accurately predicting the exact duration of a fire. In our simulations, we mitigate this issue by configuring the simulation time for CA and FlamMap (MTT) models to match the duration of the respective fire events. However, it's important to note that such precise information is typically unavailable in real forecasting scenarios. In contrast, the FIDN model doesn't require the information of fire duration, which is an important advantage for fire nowcasting.

\begin{figure}[!htb]
    \centering
    \makebox[\textwidth][c]{
    \resizebox{1.2\textwidth}{!}{%
    \begin{tabular}{cccccccc}
    \textbf{Day 2}  &
    \textbf{Final Area} &
    \multicolumn{2}{c}{\textbf{FIDN}} &
    \multicolumn{2}{c}{\textbf{CA}} &
    \multicolumn{2}{c}{\textbf{MTT}} \\
      \includegraphics[width=0.2\textwidth]{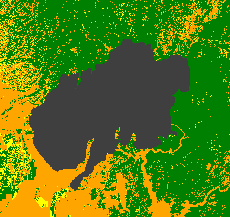} &
      \includegraphics[width=0.2\textwidth]{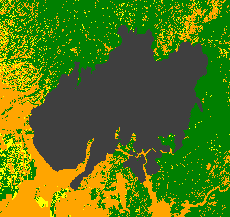} &
      \includegraphics[width=0.2\textwidth]{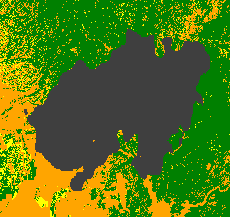} &
      \includegraphics[width=0.2\textwidth]{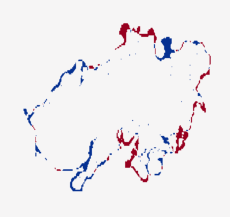} &
      \includegraphics[width=0.2\textwidth]{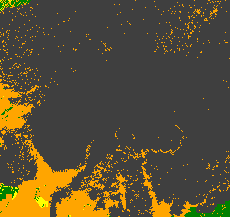} &
      \includegraphics[width=0.2\textwidth]{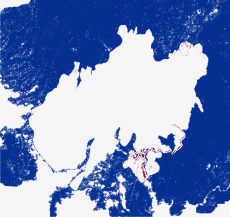} &
      \includegraphics[width=0.2\textwidth]{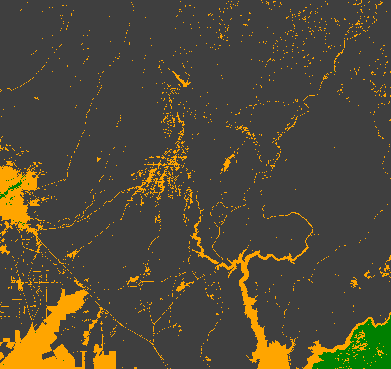} &
      \includegraphics[width=0.2\textwidth]{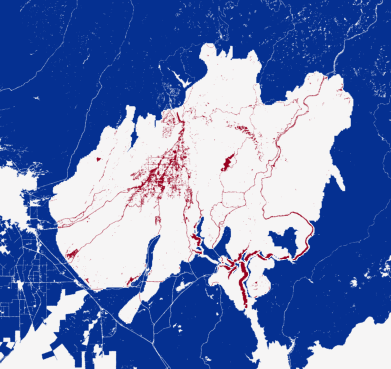} \\
      (a) origin& (b) truth & (c) predict & (d) error &
     (e) predict & (f) error & (g) predict & (h) error \\[8pt]
     \textbf{Day 2}  &
     \textbf{Final Area} &
     \multicolumn{2}{c}{\textbf{FIDN}} &
     \multicolumn{2}{c}{\textbf{CA}} &
     \multicolumn{2}{c}{\textbf{MTT}} \\
    \includegraphics[width=0.2\textwidth]{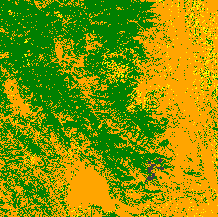} &
    \includegraphics[width=0.2\textwidth]{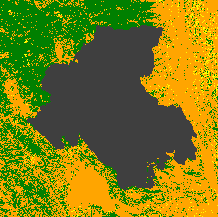} &
    \includegraphics[width=0.2\textwidth]{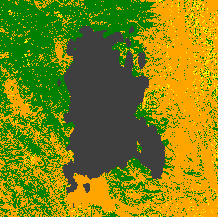} &
    \includegraphics[width=0.2\textwidth]{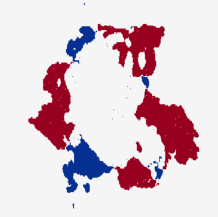} &
    \includegraphics[width=0.2\textwidth]{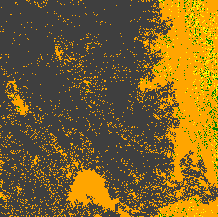} &
    \includegraphics[width=0.2\textwidth]{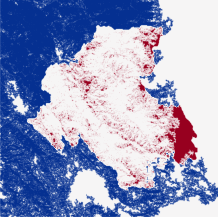} &
    \includegraphics[width=0.2\textwidth]{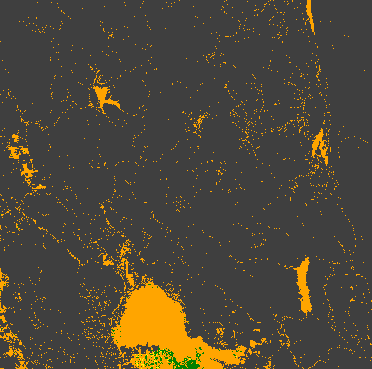} &
    \includegraphics[width=0.2\textwidth]{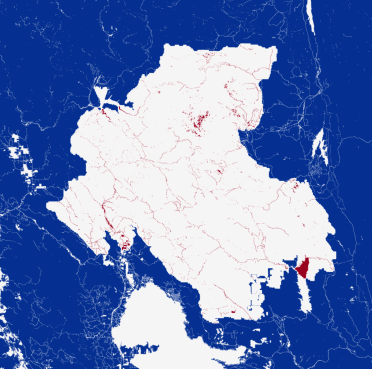} \\
    (i) origin& (j) truth & (k) predict & (l) error &
    (m) predict & (n) error & (o) predict & (p) error \\
    \end{tabular}
    }
    }
    \caption{Predicted results for the Camp(2019) and the Ranch(2018) Fire (from top to bottom)}
    \label{fig:bigfire}
\end{figure}

Figure~\ref{fig:bigfire} displays the vegetation density along with the observed and predicted burnt area of two larger fires. Figure~\ref{fig:bigfire}(a - h) shows the 2019 Camp fire event, with a burnt out area of 1178.53 $km^2$ and a duration of 13 days. As observed in Figure~\ref{fig:bigfire} (a) and (b), no significant change in terms of burnt area is observed between $\mathcal{F}^{(2)}_k$ (burnt area of day 2) and $\mathcal{F}^{(n_k)}_k$ (final burnt area). The FIDN model appears to successfully capture the pertinent influences, accurately predicting the fire front and yielding forecasts that closely align with satellite-derived observations. CA and the FlamMap (MTT) models, on the other hand, are limited to their assumptions (fires spreading outwards at every discrete time step). After 13 days of simulation, almost all of the flammable areas with high vegetation density (in green) in the ecoregion are predicted to be burnt. For the Ranch fire in Figure~\ref{fig:bigfire}(i - p) with a long fire duration of 48 days, this drawback become more remarkable.

While the prediction of the FIDN model is not flawless, it clearly provides more reasonable and interpretable predictions of the final burnt area compared to CA and FlamMap (MTT). These findings are consistent with the results in Table~\ref{table:special_fire_mertic}, where FIDN possesses substantially higher SSIM and lower MSE.

\begin{figure}[!hbt]
    \centering
    \makebox[\textwidth][c]{
    \resizebox{1.2\textwidth}{!}{%
    \begin{tabular}{cccccccc}
    \textbf{Day 2}  &
    \textbf{Final Area} &
    \multicolumn{2}{c}{\textbf{FIDN}} &
    \multicolumn{2}{c}{\textbf{CA}} &
    \multicolumn{2}{c}{\textbf{MTT}} \\
    \includegraphics[width=0.2\textwidth]{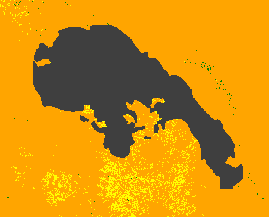} &
    \includegraphics[width=0.2\textwidth]{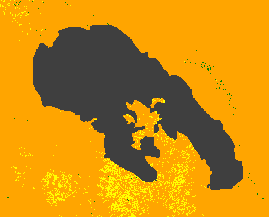} &
    \includegraphics[width=0.2\textwidth]{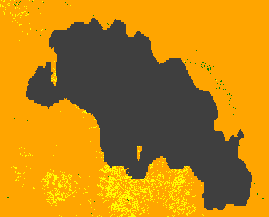} &
    \includegraphics[width=0.2\textwidth]{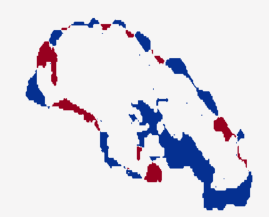} &
    \includegraphics[width=0.2\textwidth]{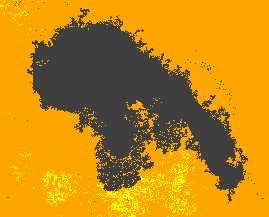} &
    \includegraphics[width=0.2\textwidth]{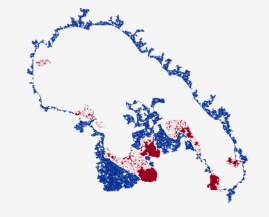} &
    \includegraphics[width=0.2\textwidth]{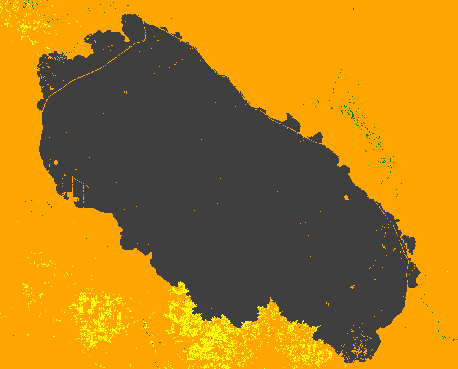} &
    \includegraphics[width=0.2\textwidth]{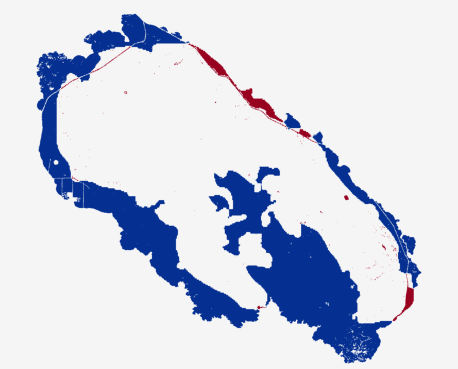} \\
    (a) origin& (b) truth & (c) predict & (d) error &
    (e) predict & (f) error & (g) predict & (h) error \\[8pt]

    \textbf{Day 2}  &
    \textbf{Final Area} &
    \multicolumn{2}{c}{\textbf{FIDN}} &
    \multicolumn{2}{c}{\textbf{CA}} &
    \multicolumn{2}{c}{\textbf{MTT}} \\
    \includegraphics[width=0.2\textwidth]{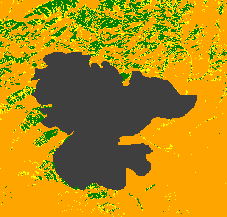} &
    \includegraphics[width=0.2\textwidth]{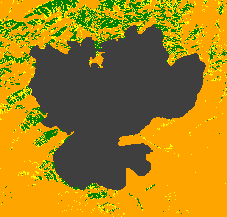} &
    \includegraphics[width=0.2\textwidth]{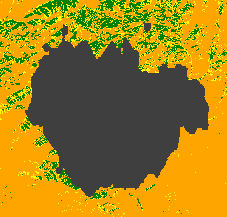} &
    \includegraphics[width=0.2\textwidth]{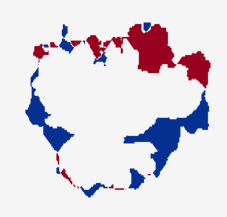} &
    \includegraphics[width=0.2\textwidth]{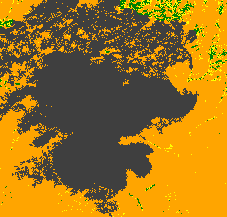} &
    \includegraphics[width=0.2\textwidth]{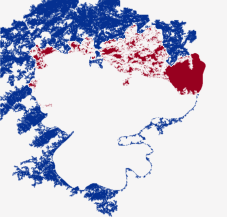} &
    \includegraphics[width=0.2\textwidth]{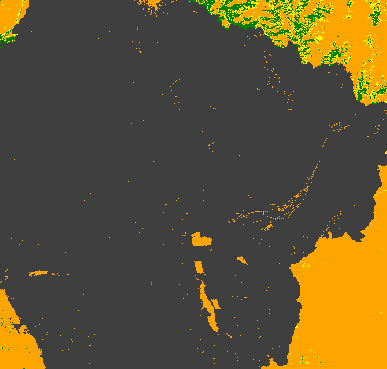} &
    \includegraphics[width=0.2\textwidth]{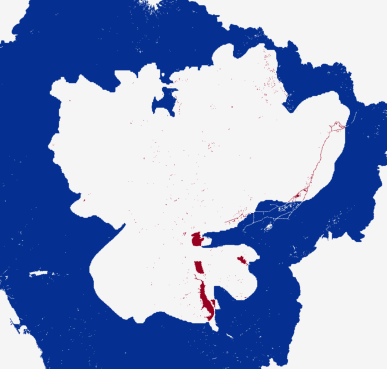} \\
    (i) origin& (j) truth & (k) predict & (l) error &
    (m) predict & (n) error & (o) predict & (p) error \\[8pt]

    \textbf{Day 2}  &
    \textbf{Final Area} &
    \multicolumn{2}{c}{\textbf{FIDN}} &
    \multicolumn{2}{c}{\textbf{CA}} &
    \multicolumn{2}{c}{\textbf{MTT}} \\
    \includegraphics[width=0.2\textwidth]{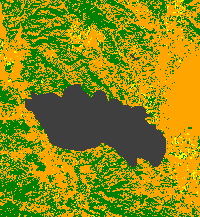} &
    \includegraphics[width=0.2\textwidth]{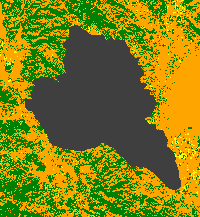} &
    \includegraphics[width=0.2\textwidth]{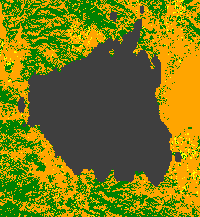} &
    \includegraphics[width=0.2\textwidth]{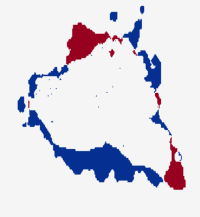} &
    \includegraphics[width=0.2\textwidth]{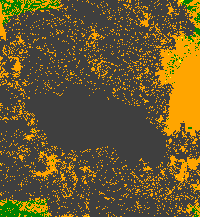} &
    \includegraphics[width=0.2\textwidth]{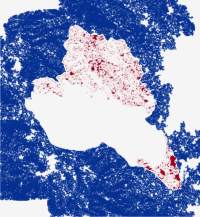} &
    \includegraphics[width=0.2\textwidth]{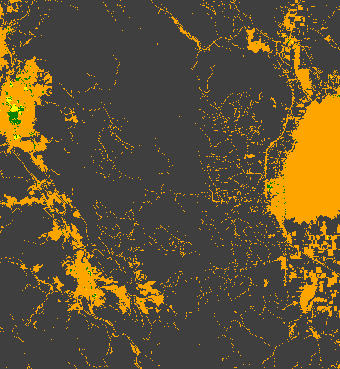} &
    \includegraphics[width=0.2\textwidth]{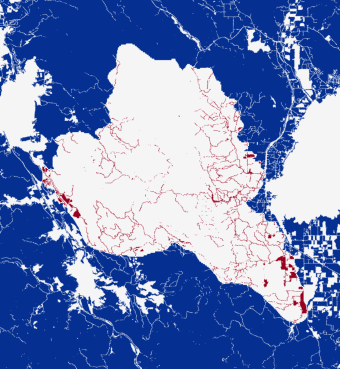} \\
    (q) origin& (r) truth & (s) predict & (t) error &
    (u) predict & (v) error & (w) predict & (x) error \\
    \end{tabular}
    }
    }
    \caption{Predicted results for the Perry(2018), the Sharps(2018) and the River(2018) Fire (from top to bottom)}
    \label{fig:midfire}
\end{figure}

Three moderate-size fire events (around 400$km^2$) are presented in Figure~\ref{fig:midfire}. FIDN demonstrates a strong understanding of the fire spread area, consistently achieving a PSNR exceeding 20 for all three fires. It's evident from the figures that the FIDN predictions closely align with the actual burnt areas. In cases of shorter-duration fires like Perry, as shown in the upper portion of Figure~\ref{fig:midfire}, CA outperforms FIDN, with a 17.8\% lower RRMSE. Additionally, MTT also provides reasonably accurate predictions. However, as the duration increases, CA and FlamMap (MTT) models clearly overestimate the final burnt area over the regions with high vegetation density, as shown in the last two fire events of Figure~\ref{fig:midfire}.

\begin{figure}[!htb]
    \centering
    \makebox[\textwidth][c]{
    \resizebox{1.2\textwidth}{!}{%
    \begin{tabular}{cccccccc}
    \textbf{Day 2}  &
    \textbf{Final Area} &
    \multicolumn{2}{c}{\textbf{FIDN}} &
    \multicolumn{2}{c}{\textbf{CA}} &
    \multicolumn{2}{c}{\textbf{MTT}} \\
      \includegraphics[width=0.2\textwidth]{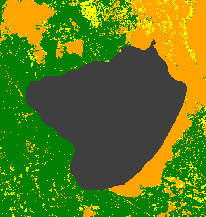} &
      \includegraphics[width=0.2\textwidth]{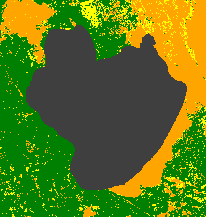} &
      \includegraphics[width=0.2\textwidth]{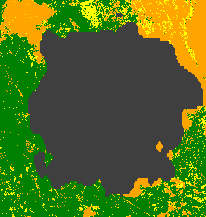} &
      \includegraphics[width=0.2\textwidth]{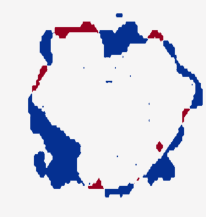} &
      \includegraphics[width=0.2\textwidth]{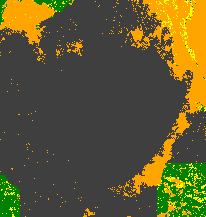} &
      \includegraphics[width=0.2\textwidth]{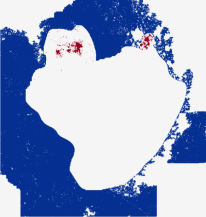} &
      \includegraphics[width=0.2\textwidth]{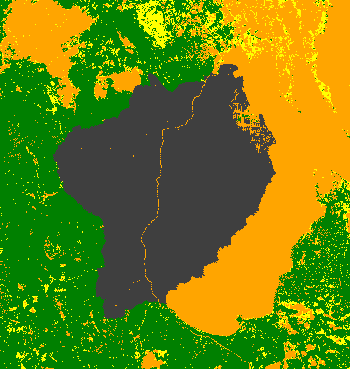} &
      \includegraphics[width=0.2\textwidth]{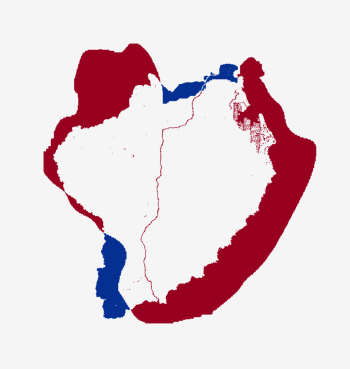}\\
      (a) origin& (b) truth & (c) predict & (d) error &
      (e) predict & (f) error & (g) predict & (h) error \\[8pt]
      \textbf{Day 2}  &
      \textbf{Final Area} &
      \multicolumn{2}{c}{\textbf{FIDN}} &
      \multicolumn{2}{c}{\textbf{CA}} &
      \multicolumn{2}{c}{\textbf{MTT}} \\

      \includegraphics[width=0.2\textwidth]{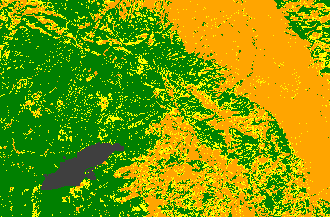} &
      \includegraphics[width=0.2\textwidth]{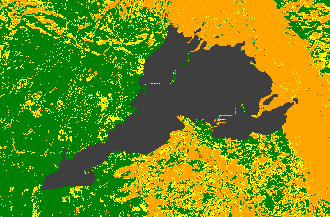} &
      \includegraphics[width=0.2\textwidth]{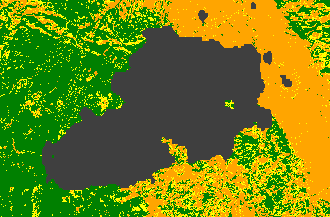} &
      \includegraphics[width=0.2\textwidth]{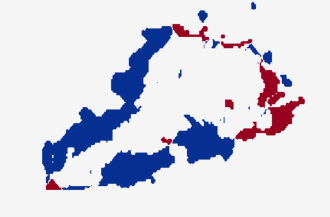} &
      \includegraphics[width=0.2\textwidth]{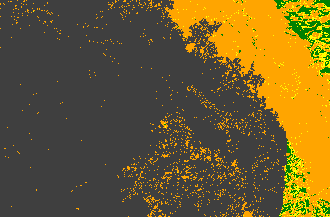} &
      \includegraphics[width=0.2\textwidth]{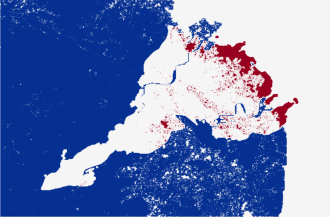} &
      \includegraphics[width=0.2\textwidth]{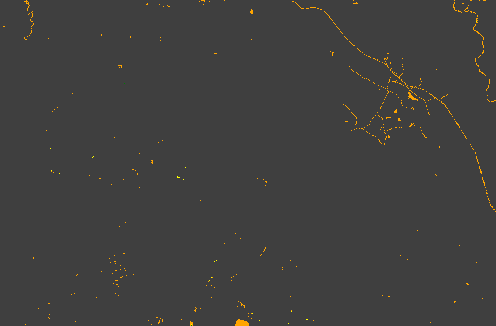} &
      \includegraphics[width=0.2\textwidth]{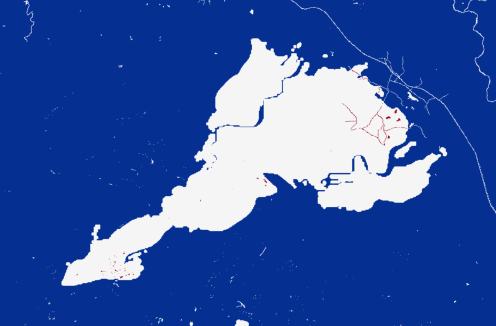}\\
      (i) origin& (j) truth & (k) predict & (l) error &
      (m) predict & (n) error & (o) predict & (p) error
    \end{tabular}
    }
    }
    \caption{Predicted results for the Whaleback(2018) and the Ryan(2018) Fire (from top to bottom)}
    \label{fig:smallfire}
\end{figure}

We find in this study that CA and FlamMap (MTT) models have more difficulties for fires with inherently small areas (less than 300$km^2$) as shown in Figure~\ref{fig:smallfire}. As observed in Figure~\ref{fig:smallfire} (e), despite the Whaleback fire having a duration of just 6 days, the CA model produces predictions with exceptionally large burnt areas due to the challenge of predicting fire duration accurately. Conversely, we make a surprising observation that the MTT algorithm provides a remarkably accurate forecast in this scenario and attains the highest SSIM score among the three models. In addition, FlamMap (MTT) completes the simulation using less than one minute. These results demonstrate the strength of the MTT algorithm in short-term predictions. However, for a longer duration fire, i.e., the Ryan fire in the bottom of Figure~\ref{fig:smallfire}, which lasted for 16 days, a substantial overestimation of MTT can still be noticed.

In summary, the analysis and demonstration presented above indicate that the current CA model and FlamMap (MTT) models are better suited for short-term predictions but have limitations in forecasting long-term fires and their resulting burnt areas. In contrast, the FIDN model not only enhances computational efficiency but, more importantly, exhibits improved model generalizability, capable of accurately predicting both relatively large and small fire events. 

\section{Conclusion and future work}
\label{section:5}

In this paper, we propose a deep learning predictive model, named Fire-Image-DenseNet (FIDN), which takes the initial burnt area (for the first three days), together with geophysical and climate data as inputs to predict the final burnt area of wildfires. We have shown that our new FIDN model produces realistic predictions of final burnt areas independent of fire size or fire duration. The structure of FIDN relies on the advanced DenseNet network that takes full advantage of convolutional neural networks and significantly reduces computational costs and computation time.

Since the model ingests remotely sensed information, it would be possible to update the predictions regularly using fire line and burn data from MODIS to take account of any changes in fire behaviour. At present, the model utilizes a combination of current reanalysis data. Future research will explore the integration of real-time data sources to enable real-time forecasting capabilities. This would make it possible to use the model to determine the potential impact of specific fire-fighting strategies to manage ongoing wildfire events, such as the optimal location for the application of fire retardants or the creation of fire breaks. While it would be useful to test the FIDN model in other regions, the method is data-agnostic and could be applied to wildfires in other areas globally. Thus, the FIDN model can provide a useful tool to enable land managers and fire services to deal with wildfires more promptly, thus reducing the negative impacts of fire on the environment. 

\section*{Acknowledgements}
This research is funded by the Leverhulme Centre for Wildfires, Environment and Society through the Leverhulme Trust, grant number RC-2018-023. 

\section*{Code and data availability}
The codes that were used for building and testing the FIDN model using Python language (3.7) are available at 
\url{https://github.com/DL-WG/FIDN}.

\section*{CRediT authorship contribution statement}

\noindent B.Pang: Methodology, Software, Formal analysis,
Writing - Original draft preparation. \\
S.Cheng: Conceptualization, Methodology, Software, Supervision,
Writing - Original draft preparation.\\
Y.Huang: Data curation, Writing - Review \& Editing \\
Y.Jin: Data curation, Writing - Review \& Editing. \\
C.Prentice: Supervision, Writing - Review \& Editing. \\
Y.Guo: Funding acquisition, Writing - Review \& Editing\\
S.Harrison: Supervision, Writing - Review \& Editing. \\
R.Arcucci: Conceptualization, Supervision, Writing - Review \& Editing. \\

\section*{Appendix: comparison against an autoregressive fire prediction model}

In our study, we initially evaluated various deep learning models and ultimately selected DenseNet for our Fire-Image-DenseNet (FIDN) approach. To explore alternative models further, we considered ConvLSTM, a state-of-the-art deep learning model commonly utilized for forecasting spatial-temporal sequences. However, the primary objective of our research is to predict the final burnt area of wildfires from the onset of ignition. ConvLSTM, by contrast, is inherently designed to predict subsequent frames within a sequence, which poses limitations in our context due to the variable durations of wildfires. 

Subsequently, we analyzed and compared the performance of the two models in predicting the final burned area of wildfires based on the data presented in Table \ref{table:appendix_avg_mertic}. The results for SSIM indicate that both the mean and median values for FIDN are significantly higher than those for ConvLSTM, suggesting that FIDN performs better in capturing the structure and details of the images. Similarly, the PSNR results, while showing higher values for ConvLSTM, also exhibit a larger standard deviation, indicating instability and unreliability in its predictions. In terms of MSE, while ConvLSTM shows lower values, our images are largely binary, with most pixels indicating unburned areas (value 0), making MSE less reflective of accuracy in the actually burned regions. The RRMSE metric, on the other hand, reveals that FIDN achieves lower relative error, emphasizing its ability to accurately predict the crucial non-zero areas that define the final burnt regions in these binary images. Notably, in the RRMSE metric, both the mean and median values for FIDN considerably outperform those of ConvLSTM, underscoring its advantage in relative error measurement. Our tests showed ConvLSTM could only predict the fourth day’s burned area accurately, leading to predictions that closely resemble the initial input image rather than the final burnt area.

\begin{table}[hbp]
    \centering
    \resizebox{0.8\textwidth}{!}{
    \begin{tabular}{@{}llllll@{}}
    \toprule
     &  & \textbf{FIDN} & \textbf{ConvLSTM} \\ \midrule
    \multirow{2}{*}{\textbf{SSIM}} & Mean ± stddev & 0.971 ± 0.015 & 0.695 ± 0.012 \\
     & Median (IQR) &0.974 (0.015) & 0.701 (0.013) \\ \midrule
    \multirow{2}{*}{\textbf{PSNR}} & Mean ± stddev & 20.993 ± 2.787 & 25.319 ± 6.442 \\
     & Median (IQR) & 21.473 (3.638) & 26.650 (10.957) \\ \midrule
    \multirow{2}{*}{\textbf{MSE}} & Mean ± stddev & 0.010 ± 0.008 & 0.008 ± 0.013 \\
     & Median (IQR) & 0.007 (0.006) & 0.002 (0.009) \\ \midrule
    \multirow{2}{*}{\textbf{RRMSE}} & Mean ± stddev & 0.825 ± 0.212 & 1.252 ± 1.337 \\
     & Median (IQR) & 0.821 (0.267) & 0.715 (1.104) \\ \bottomrule
    \end{tabular}
    }
    \caption{Performance statistics (mean, median and standard deviation) of model predictions of final burnt area summarized over 23 wildfires from the test dataset. The performance of the Fire-Image-DenseNet (FIDN) model is compared to predictions of the ConvLSTM model. The five evaluation metrics are included: the structural similarity index measure (SSIM), a normalised value between 1 for perfect correlation and 0 for no correlation; the peak signal-to-noise ratio (PSNR); the mean square error (MSE) in $km^2$ and the relative root mean square error (RRMSE).}
    \label{table:appendix_avg_mertic}
\end{table}

\begin{figure}[!htb]
    \centering
    \makebox[\textwidth][c]{
    \resizebox{1\textwidth}{!}{%
    \begin{tabular}{cccccc}
     \textbf{Day 2}  &
     \textbf{Final Area} &
     \multicolumn{2}{c}{\textbf{FIDN}} &
     \multicolumn{2}{c}{\textbf{ConvLSTM}} \\
    \includegraphics[width=0.2\textwidth]{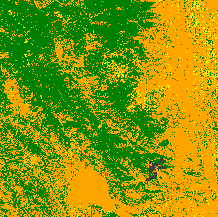} &
    \includegraphics[width=0.2\textwidth]{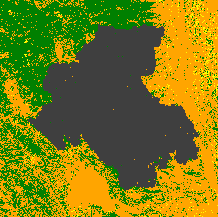} &
    \includegraphics[width=0.2\textwidth]{dense_1_Ranch_2018_NAT_predict.png} &
    \includegraphics[width=0.2\textwidth]{dense_1_Ranch_2018_NAT_error.png} &
    \includegraphics[width=0.2\textwidth]{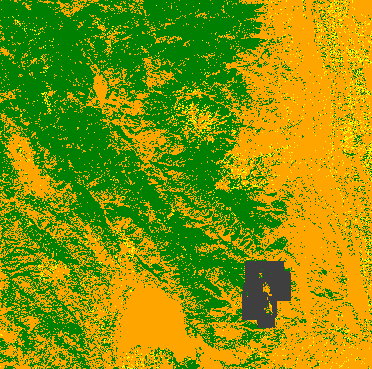} &
    \includegraphics[width=0.2\textwidth]{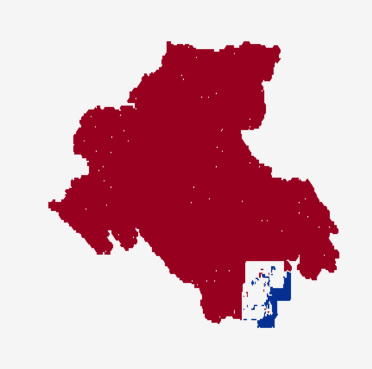} \\
    (g) origin& (h) truth & (i) predict & (j) error &
    (k) predict & (l) error \\[8pt]
    \textbf{Day 2}  &
    \textbf{Final Area} &
    \multicolumn{2}{c}{\textbf{FIDN}} &
    \multicolumn{2}{c}{\textbf{ConvLSTM}} \\
      \includegraphics[width=0.2\textwidth]{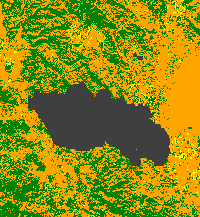} &
      \includegraphics[width=0.2\textwidth]{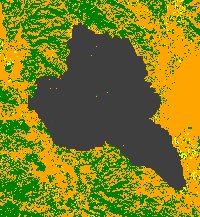} &
      \includegraphics[width=0.2\textwidth]{dense_3_River_2018_NAT_predict.png} &
      \includegraphics[width=0.2\textwidth]{dense_3_River_2018_NAT_error.png} &
      \includegraphics[width=0.2\textwidth]{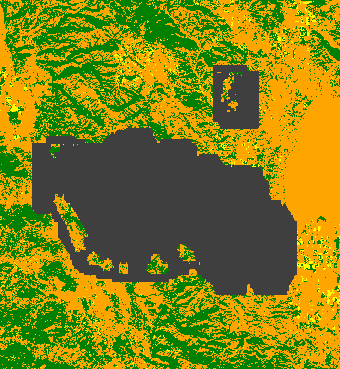} &
      \includegraphics[width=0.2\textwidth]{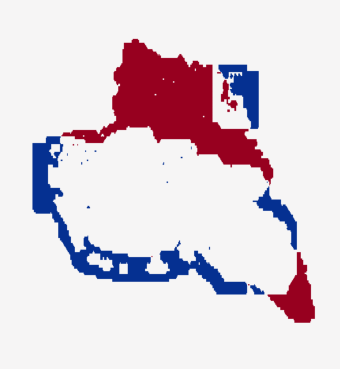} \\
      (a) origin& (b) truth & (c) predict & (d) error &
      (e) predict & (f) error \\[8pt]
    \textbf{Day 2}  &
    \textbf{Final Area} &
    \multicolumn{2}{c}{\textbf{FIDN}} &
    \multicolumn{2}{c}{\textbf{ConvLSTM}} \\
      \includegraphics[width=0.2\textwidth]{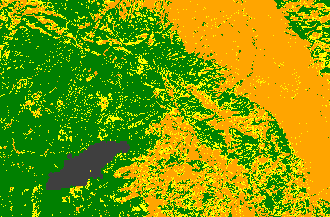} &
      \includegraphics[width=0.2\textwidth]{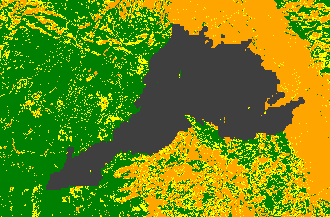} &
      \includegraphics[width=0.2\textwidth]{dense_5_Ryan_2018_NAT_predict.png} &
      \includegraphics[width=0.2\textwidth]{dense_5_Ryan_2018_NAT_error.png} &
      \includegraphics[width=0.2\textwidth]{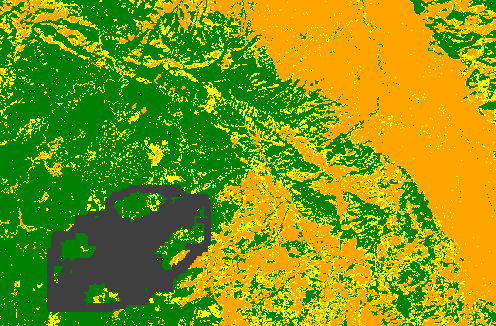} &
      \includegraphics[width=0.2\textwidth]{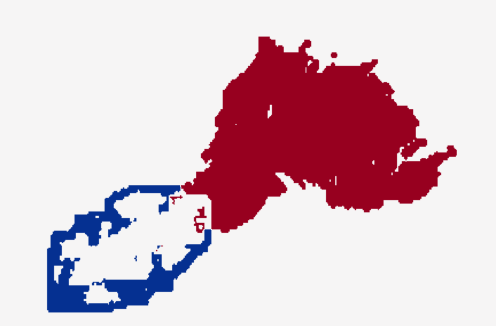} \\
      (a) origin& (b) truth & (c) predict & (d) error &
      (e) predict & (f) error \\[8pt]
    \end{tabular}
    }
    }
    \caption{Predicted results for the Ranch(2018), the River(2018), and the Ryan(2018) Fire (from top to bottom)}
    \label{fig:appendix_bigfire}
\end{figure}

To further demonstrate FIDN’s predictive accuracy for final burned areas, we selected three prolonged wildfires for case analysis. These images illustrated the significant discrepancies between the predictions made by the ConvLSTM model and the actual burned areas (See Figure \ref{fig:appendix_bigfire}). Through these real-world examples, it becomes evident that while ConvLSTM may perform adequately in certain time series prediction tasks, its predictive capabilities fall short in the complex and dynamically evolving context of wildfires, thereby validating our previous claims.

\clearpage
\footnotesize
\bibliographystyle{plainnat}
 \bibliography{cas-refs}






\end{document}